\documentclass{article} % For LaTeX2e
\usepackage{iclr2024_conference,times}

\usepackage{amsmath,amsfonts,bm}

\def\eqref#1{equation~\ref{#1}}
\def\1{\bm{1}}

\DeclareMathAlphabet{\mathsfit}{\encodingdefault}{\sfdefault}{m}{sl}
\SetMathAlphabet{\mathsfit}{bold}{\encodingdefault}{\sfdefault}{bx}{n}

\usepackage{graphicx}
\usepackage{float}
\usepackage{subfig}
\usepackage{hyperref}
\usepackage{url}
\usepackage{threeparttable}
\usepackage{algorithm}
\usepackage{algorithmic}
\usepackage{booktabs}
\usepackage{multirow, array}  
\usepackage{wrapfig}
\usepackage{pifont} 
\usepackage{diagbox} 
\usepackage{arydshln}
\usepackage{graphicx}
\usepackage{enumitem}
\usepackage{amsmath}

\newlength\savewidth\newcommand\shline{\noalign{\global\savewidth\arrayrulewidth
  \global\arrayrulewidth 1pt}\hline\noalign{\global\arrayrulewidth\savewidth}}
  
\newcommand{\cmark}{\ding{51}}

\title{MetaCoCo: A New Few-Shot Classification Benchmark with Spurious Correlation}

\author{Min Zhang$^{1}$\qquad Haoxuan Li$^{2}$\qquad Fei Wu$^{1}$\qquad Kun Kuang$^{1}$\thanks{Corresponding author.}\\
{$^1$Zhejiang University} \quad {$^2$Peking University}\\
{\tt\small \{zhangmin.milab, wufei, kunkuang\}@zju.edu.cn}, 
{\tt\small hxli@stu.pku.edu.cn}
}

\iclrfinalcopy % Uncomment for camera-ready version, but NOT for submission.

\begin{document}

\maketitle

\begin{abstract}
    Out-of-distribution (OOD) problems in few-shot classification (FSC) occur when novel classes sampled from testing distributions differ from base classes drawn from training distributions, which considerably degrades the performance of deep learning models deployed in real-world applications. Recent studies suggest that the OOD problems in FSC mainly including: (a) cross-domain few-shot classification (CD-FSC) and (b) spurious-correlation few-shot classification (SC-FSC). Specifically, CD-FSC occurs when a classifier learns transferring knowledge from base classes drawn from \underline{seen} training distributions but recognizes novel classes sampled from \underline{unseen} testing distributions. In contrast, SC-FSC arises when a classifier relies on \underline{non-causal} features (or contexts) that happen to be \underline{correlated} with the labels (or concepts) in base classes but such relationships no longer hold during the model deployment. Despite CD-FSC has been extensively studied, SC-FSC remains understudied due to lack of the corresponding evaluation benchmarks. To this end, we present \textbf{Meta} \textbf{Co}ncept \textbf{Co}ntext (MetaCoCo), a benchmark with spurious-correlation shifts collected from real-world scenarios. Moreover, to quantify the extent of spurious-correlation shifts of the presented MetaCoCo, we further propose a metric by using CLIP as a pre-trained vision-language model. Extensive experiments on the proposed benchmark are performed to evaluate the state-of-the-art methods in FSC, cross-domain shifts, and self-supervised learning. The experimental results show that the performance of the existing methods degrades significantly in the presence of spurious-correlation shifts. We open-source all codes of our benchmark and hope that the proposed MetaCoCo can facilitate future research on spurious-correlation shifts problems in FSC. The code is available at: \href{https://github.com/remiMZ/MetaCoCo-ICLR24}{https://github.com/remiMZ/MetaCoCo-ICLR24}.
\end{abstract}
\section{Introduction}
\label{sec:intro}
\begin{figure}[tbp]
    \centering
    \subfloat[Cross-domain few-shot classification (CD-FSC)]{
    \begin{minipage}[t]{0.99\linewidth}
        \centering
        \includegraphics[scale = 0.465]{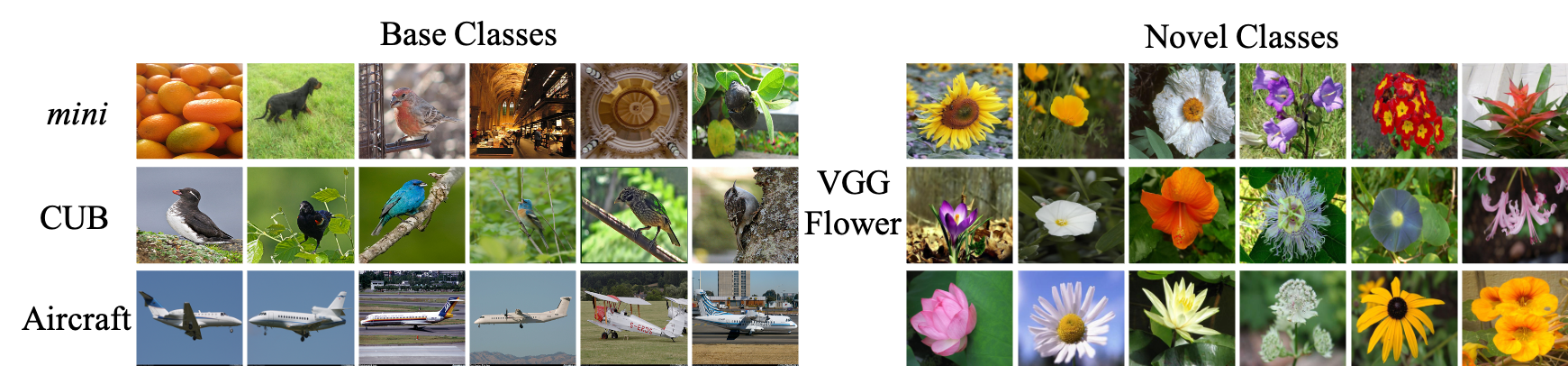}
    \end{minipage}
    }
    \vspace{-1mm}
    \subfloat[Spurious-correlation few-shot classification (SC-FSC)]{
    \begin{minipage}[t]{0.99\linewidth}
        \centering
        \includegraphics[scale = 0.465]{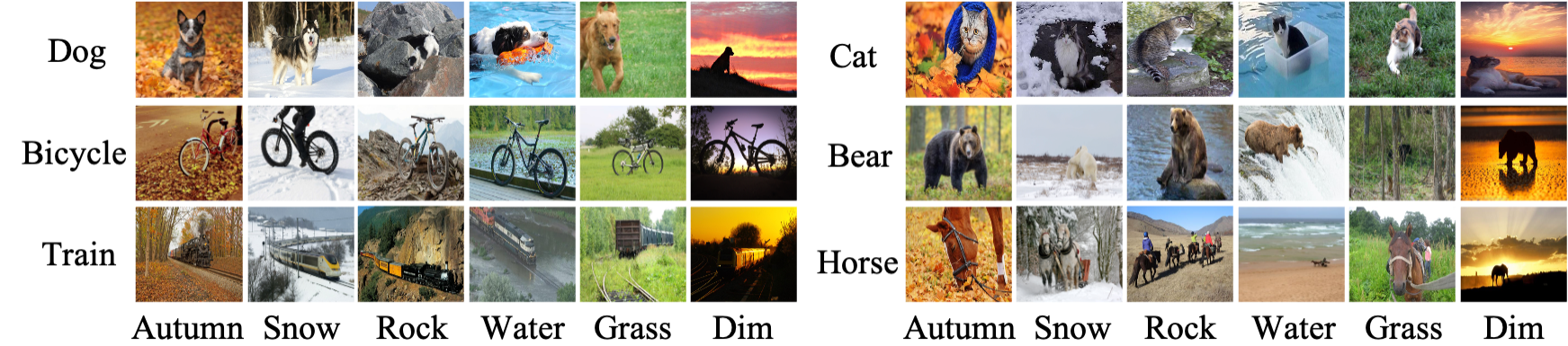}
    \end{minipage}
    }
\vspace{-2mm}
\caption{Example of cross-domain shifts and spurious-correlation shifts in FSC. (a) In {Meta-dataset} with cross-domain shifts~\citep{triantafillou2019meta}, the model is trained on base classes sampled from three datasets including \textit{mini}ImageNet, CUB-200-2011 and Aircraft, then tested on novel classes drawn from VGG Flower. (b) In our proposed MetaCoCo with spurious-correlation shifts, each class (or concept, \textit{e.g.}, dog) consists of different backgrounds (or context, \textit{e.g.}, autumn).}
\label{fig:motivation}
\vspace{-4mm}
\end{figure}

Few-shot classification (FSC) aims to recognize unlabeled images (or query sets) from novel classes with only a few labeled images (or support sets) by transferring knowledge learned from base classes. Despite the impressive advances in the FSC, in real-world applications, out-of-distribution (OOD) problems in FSC occur when the novel classes sampled from testing distributions differ from the base classes drawn from training distributions, which significantly degrades the performance and robustness of deep learning models, and has gained increasing attention in recent years~\citep{song2022comprehensive,li2021libfewshot}. As shown in Figure~\ref{fig:motivation}, the OOD problems in FSC can be broadly categorized into two categories with different forms of distribution shifts: (a) cross-domain few-shot classification (CD-FSC) and (b) spurious-correlation few-shot classification (SC-FSC), as established by previous works~\citep{triantafillou2019meta,YueZS020,luo2021rectifying,li2022cross}.
  
\textbf{Cross-domain few-shot classification (CD-FSC).}
Cross-domain shifts occur when a classifier learns transferring knowledge from base classes drawn from \underline{seen} training distributions but recognizes novel classes sampled from \underline{unseen} testing distributions. For example, in COVID-19 predictions, we may want to train a model on patients from a few sampled countries and then deploy the trained model to a broader set of countries. Existing OOD methods in FSC have shown considerable progress in solving the cross-domain shifts problem~\citep{hou2019cross,doersch2020crosstransformers,GuoCKCSSRF20,WangD21,sun2021explanation,liang2021boosting,WangD21,li2022tdr,li2022stabledr,oh2022understanding,zhang2020knowledge,zhang2022domain}. Meanwhile, two standard cross-domain benchmarks have been proposed to evaluate the effectiveness of these methods, \textit{i.e.}, {Meta-dataset}~\citep{triantafillou2019meta} consisting of 10 existing datasets, and BSCD-FSL~\citep{GuoCKCSSRF20} consisting of 4 existing datasets. Figure~\ref{fig:motivation}(a) shows the example of cross-domain shifts on {Meta-dataset}, where \textit{mini} (\textit{mini}ImageNet), CUB (CUB-200-2011) and Aircraft are used as the base classes with VGG Flower as the novel classes, with each dataset exhibits a distinct distribution.

\textbf{Spurious-correlation few-shot classification (SC-FSC).}
Spurious-correlation shifts arise when a classifier relies on spurious, non-causal context features that are not essential to the true label or concept, which can significantly reduce the robustness and generalization ability of the model. 
In the COVID-19 example, a recent nationwide cross-sectional study found spurious correlations between long-term PM$_{2.5}$ exposure and COVID-19 deaths in the United States due to county-level socioeconomic and demographic variables as confounders~\citep{wu2020exposure}.
% a model trained on the entire population of a country may associate the labels with government policies or demographic characteristics (\textit{e.g.}, gender and age), causing the model to fail on other countries when such a relationship does not hold in reality~\citep{yao2022improving}. 
To this end, models trained on base classes with spurious features and evaluated on novel classes without the relationship suffer substantial drops in performance. 
As shown in Figure~\ref{fig:motivation}(b), we show the example of spurious-correlation shifts in our proposed benchmark, where each class presents a range of non-causal contexts, such as autumn or snow. %Moreover, base classes and novel classes have some non-overlapping contexts, highlighting the presence of distribution shifts. 
Meanwhile, the concepts of the base classes and the novel classes would be distinct in the FSC problem, \textit{e.g.}, ``dog in the autumn" in the base class and ``cat in the autumn" in the novel class, which emphasizes the impact of spurious correlation between concepts and contexts in the proposed benchmark.
Despite the widespread of spurious-correlation shifts in the real-world FSC problems~\citep{wang2017community,YueZS020,luo2021rectifying,zhang2023map}, SC-FSC remains understudied due to lack of the corresponding evaluation benchmarks.

% \begin{figure}[tbp]
%     \centering
% \includegraphics[width=0.99\textwidth]{figures/correlation.pdf}
%     \vspace{-7.5mm}
%     \caption{Example of cross-domain shifts and spurious-correlation shifts. In cross-domain shifts with \textbf{Meta-dataset} (a), the model is trained on base classes sampled from three datasets including \textit{mini} (\textit{mini}ImageNet), CUB (CUB-200-2011) and Aircraft, then tested on novel classes drawn from VGG Flower. In spurious-correlation shifts with our proposed MetaCoCo (b), each class (or concept) (\textit{e.g.}, dog) consists of different backgrounds (or context) (\textit{e.g.}, autumn or grass).}
%     \label{fig:motivation}
%     \vspace{-5mm}
% \end{figure}

\textbf{Shortcomings of spurious-correlation shifts benchmarks in traditional machine learning.} 
Recently, spurious-correlation shifts in traditional machine learning (TML) have been investigated extensively~\citep{arjovsky2019,sagawa2019,rosenfeld2020risks,ahmed2020systematic,bae2021meta,shen2021towards}, and various benchmarks have been created, including toy datasets, \textit{e.g.}, ColoredMNIST~\citep{arjovsky2019}, and real-world datasets, \textit{e.g.}, NICO~\citep{he2021towards}. These TML benchmarks cannot be used directly to evaluate the performance in FSC problems with spurious-correlation shifts, following the reasons below: 
(1) \textbf{The number of classes.} Most TML benchmarks are the binary classification problem, but for FSC problems, we need enough classes to split base and novel classes.
(2) \textbf{The number of samples.} FSC needs adequate samples from base classes to learn the transferring knowledge to novel classes with a few labeled images. 
(3) \textbf{The number of contexts.} Contexts in TML benchmarks are commonly limited, but FSC with many classes requires more contexts to build stronger spurious-correlation shifts. To the best of our knowledge, there does not exist a unified study and the benchmark of spurious-correlation shifts for FSC. 

In this paper, we present \textbf{Meta} \textbf{Co}ncept \textbf{Co}ntext (MetaCoCo), a large-scale benchmark with a total of 175,637 images, 155 contexts and 100 classes, with spurious-correlation shifts arising from various contexts in the real-world scenarios. The basic idea of constructing spurious-correlation shifts is to label the images with the main concepts and contexts. For example, in the category with ``dog" as the main concept, the images are categorized into different contexts such as ``autumn", ``snow", and ``rock", which denotes that the ``dog" is in the autumn, in the snow, or on the rock, respectively. With the help of these contexts, one can easily design a spurious-correlation-shift setting by training the model in some contexts and testing the model in other unseen contexts for studying spurious-correlation shifts as well as the unseen concepts for studying few-shot classification problems.

Furthermore, we propose a metric by using CLIP as a pre-trained vision-language model to quantify and compare the extent of spurious correlations on MetaCoCo and other FSC benchmarks. We conduct extensive experiments on MetaCoCo to evaluate the state-of-the-art methods in FSC, cross-domain shifts, and self-supervised learning. We open-source
all codes for our benchmark and hope the proposed MetaCoCo will facilitate the development of spurious-correlation robust models.
\section{Comparison with Existing Benchmarks} 
\label{sec:rela}

MetaCoCo provides a unified framework to facilitate the development of models robust to spurious-correlation shifts in FSC. We next discuss how MetaCoCo is related to existing benchmarks.

\textbf{Relation to few-shot classification benchmarks.}
Few-shot classification (FSC) has attracted attention for its ability to recognize novel classes using few labeled images.
Many methods have been proposed to solve the FSC problems, including (1) \textit{Fine-tuning based methods}~\citep{ChenLKWH19,tian2020rethinking,chen2021meta}, which address the problem by \textit{learn to transfer}. 
% They use a pre-training strategy to learn the knowledge from base classes during training and transfer the learned knowledge to fast adapt novel classes with only few labeled images during testing. 
(2) \textit{Metric-based methods}~\citep{VinyalsBLKW16,snell2017prototypical,li2019revisiting,zhang2022tree}, which solve the problem by \textit{learn to compare}. 
% They compare the similarity between the support and query set based on a metric space such as Euclidean or cosine distance. 
(3) \textit{Meta-based methods}~\citep{FinnAL17,RusuRSVPOH19,bae2021meta,zhang2020knowledge}, which tackle the problem by \textit{learn to learn}. 
% They learn the good model initialization parameters through the bi-level optimization, which aims to fast adapt to novel classes with a few steps of gradient updates by using only few labeled images. 

Many FSC benchmarks have been proposed to evaluate the effectiveness of these methods, including \textit{mini}ImageNet~\citep{VinyalsBLKW16}, Places~\citep{zhou2017places}, CIFAR-FS~\citep{bertinetto2018meta}, Plantae~\citep{van2018inaturalist}, CUB-200-2011~\citep{wah2011caltech}, Stanford Dogs~\citep{khosla2011novel}, Stanford Cars~\citep{krause20133d}, etc. These datasets are generally divided into training, validation and testing sets with non-overlap classes. While these datasets are useful testbeds for verifying FSC methods, they follow the independent and identically distributed (IID) assumption. 

\textbf{Relation to cross-domain shifts FSC benchmarks.}
Cross-domain shifts have been widely studied in the FSC community, which aims to learn the transferring knowledge from seen training distributions to recognize unseen testing distributions. Many CD-FSC methods have been proposed to address the cross-domain problem~\citep{TsengLH020,sun2021explanation,liang2021boosting,WangD21,li2022cross,zhang2022domain}, which can be mainly divided into bi-level optimization~\citep{TsengLH020,triantafillou2021learning,li2023balancing,zhang2023rotogbml}, domain adversarial learning~\citep{motiian2017few,zhao2021domain}, adversarial data augmentation~\citep{WangD21,sun2021explanation}, and module modulation~\citep{liu2021multi,li2022cross}. 
Some benchmarks have been proposed to evaluate the effectiveness of these CD-FSC methods, including Meta-dataset~\citep{triantafillou2019meta} consisting of 10 existing datasets, and BSCD-FSL~\citep{GuoCKCSSRF20} consisting of 4 existing datasets. 
They usually use the leave-one-domain-out setting as the testing domain and the others as training domains. However, these benchmarks use different datasets as domains to construct cross-domain distribution shifts, causing them to fail to reflect spurious correlation shifts that occur in real-world applications 
{(see more discussion in Appendix~\ref{app-a})}.
% {(see more discussion in Appendix A)}.

\textbf{Relation to spurious-correlation shifts TML benchmarks.}
Spurious-correlation shifts have been studied recently in traditional machine learning (TML)~\citep{sagawa2019,krueger2021out,yao2022improving,bai2024prompt,tang2024modelgpt}. Many methods mainly focus on causal learning~\citep{peters2015causal,kuang2018stable,kamath2021does,Wu-etal2022-framework,wang2024optimal,li2024kernel,zhu2024contrastive}, invariant learning~\citep{arjovsky2019,chang2020invariant,rosenfeld2020risks,huang2023pareto}, and distributionally robust optimization~\citep{arjovsky2019}, etc.
Some toy benchmarks, \textit{e.g.}, ColoredMNIST~\citep{arjovsky2019} and real-world benchmarks, \textit{e.g.}, NICO~\citep{he2021towards} and MetaShift~\citep{liang2022metashift}, have been proposed to evaluate the performance of these methods.
These TML benchmarks do not be used directly in the FSC setting, due to lack of sufficient classes, number of samples, and number of contexts.
Although IFSL~\citep{YueZS020} and COSOC~\citep{luo2021rectifying} have experimentally proved the importance of spurious-correlation shifts, there is still a lack of a benchmark for evaluation. Therefore, we propose MetaCoCo in this paper to reflect spurious-correlation shifts arising in real-world scenarios.
\section{Problem and Evaluation Settings}
\label{sec:problem}
FSC aims to recognize unlabeled images (or query sets) from novel classes with only few labeled images (or support sets). Following the previous studies~\citep{VinyalsBLKW16,tian2020prior}, we adopt an episodic paradigm to train and evaluate the few-shot models. Specifically, each $N$-way $K$-shot episode $\mathcal{T}_{e}$ has a support set $\mathcal{S}_{e}=\{(x_{i}, y_{i}): i=1, \dots, I_s\}$ and a query set $\mathcal{Q}_{e}=\{(x_{i}, y_{i}): i=I_{s}+1, \dots, I_{s}+I_{q}\}$, where $x_{i} \in \mathcal{X}$ is the image and $y_{i} \in \mathcal{Y}$ is the label from a set of $N$ classes $\mathcal{C}_{e}$, with $I_{s}=N \cdot K$ and $I_{q}$ be the image numbers in the support and query set, respectively.

Let $\mathcal{S}_{e}(\mathcal{X})$ and $\mathcal{Q}_{e}(\mathcal{X})$ be the image spaces of $\mathcal{S}_{e}$ and $\mathcal{Q}_{e}$, and $\mathcal{S}_{e}(\mathcal{Y})$ and $\mathcal{Q}_{e}(\mathcal{Y})$ be the corresponding label spaces, respectively. The label space of $\mathcal{S}_{e}$ and $\mathcal{Q}_{e}$ is same but the image space is different, \textit{i.e.}, $\mathcal{S}_{e}(\mathcal{X}) \neq \mathcal{Q}_{e}(\mathcal{X})$ and $\mathcal{S}_{e}(\mathcal{Y}) = \mathcal{Q}_{e}(\mathcal{Y})$. 
During the training phase, for meta-based and metric-based methods, episodes are randomly sampled from the base classes set $\mathcal{D}_{b}$ to train the model. Instead, for fine-tuning based methods, a mini-batch images is randomly sampled from $\mathcal{D}_{b}$ to train the model.
During the testing phase, the trained model is fine tuned with $\mathcal{S}_{e}$ and evaluated with $\mathcal{Q}_{e}$ in novel episodes sampled from the novel classes set $\mathcal{D}_{n}$.
Note that $\mathcal{D}_{b}$ contains more images and classes compared with $\mathcal{D}_{n}$ but label spaces are disjoint, 
\textit{i.e.}, $\mathcal{D}_{b}(\mathcal{Y}) \neq  \mathcal{D}_{n}(\mathcal{Y})$\footnote{$\mathcal{D}_{b}(\mathcal{Y})$ and  $\mathcal{D}_{n}(\mathcal{Y})$ can be defined similarly, meaning the label spaces of $\mathcal{D}_{b}$ and $\mathcal{D}_{n}$, respectively.}.
%Distributions of $\mathcal{D}_{b}$ and $\mathcal{D}_{n}$ are assumed to be independently and identically distribution (IID). 
The model architectures have a feature encoder $f_{\theta}$ and a classifier $c_{\phi}$ parameterized by $\theta$ and $\phi$. The $f_{\theta}$ aims to extract features, $f_{\theta}: \mathcal{X}\rightarrow \mathcal{Z}$, and the $c_{\phi}$ predicts the class of extracted features, $c_{\phi}: \mathcal{Z}\rightarrow \mathcal{Y}$. 

\subsection{Cross-Domain Shifts and Spurious-Correlation Shifts}
\label{subsec:shifts}

\begin{table}[tbp]
    \centering
    \caption{A summary of the existing benchmarks and our proposed spurious-correlation benchmark, \textit{i.e.}, MetaCoCo. $\mathcal{C}$ and $\mathcal{N}$ are the number of classes and samples, respectively. The subscripts ``all", ``train", ``val" and ``test" mean the all dataset, training set, validation, and testing set, respectively.}
    \setlength{\tabcolsep}{2.5pt}
    \vspace{-3mm}
    \resizebox{1.0\textwidth}{!}{
        \begin{tabular}{l|cccc|cccc|c|c}
            \shline
            Dataset & $\mathcal{C}_{all}$ & $\mathcal{C}_{train}$ & $\mathcal{C}_{val}$ & $\mathcal{C}_{test}$ & $\mathcal{N}_{all}$ & $\mathcal{N}_{train}$ & $\mathcal{N}_{val}$ & $\mathcal{N}_{test}$ & Context & Similarity \\ 
            \hline
            \textit{mini}ImageNet~\citep{VinyalsBLKW16} & 100 & 64 & 16 & 20 & 60,000 & 38,400 & 9,600 & 12,000 & 0 & 0.211 \\
            CIFAR-FS~\citep{krizhevsky2009learning} & 100 & 64 & 16 & 20 & 60,000 & 38,400 & 9,600 & 12,000 & 0 & 0.181 \\
            Stanford Dogs~\citep{khosla2011novel} & 120 & 70 & 20 & 30 & 20,580 & 12,165 & 3,312 & 5,103 & 0 & 0.244 \\
            Stanford Cars~\citep{krause20133d} & 196 & 130 & 17 & 49 & 16,185 & 10,766 & 1,394 & 4,025 & 0 & 0.164 \\
            Aircraft~\citep{wah2011caltech} & 100 & 70 & 15 & 15 & 10,000 & 5,000 & 2,500 & 2,500 & 0 & 0.228 \\
            CUB-200-2011~\citep{wah2011caltech} & 200 & 140 & 30 & 30 & 11,788 & 7,648 & 1,182 & 2,958 & 0 & 0.266 \\
            Describable Textures~\citep{cimpoi2014describing} & 47 & 33 & 7 & 7 & 5,640 & 3,960 & 840 & 840 & 0 & 0.194 \\
            Traffic Signs~\citep{houben2013detection} & 43 & - & - & 43 & 50,000 & - & - & 50,000 & 0 & 0.193 \\
            Omniglot~\citep{lake2015human} & 50 & 25 & 5 & 20 & 32,460 & 17,660 & 1,620 & 13,180 & 0 & 0.212 \\
            Fungi~\citep{schroeder2018fgvcx} & 1394 & 994 & 200 & 200 & 89,760 & 64,449 & 12,195 & 13,116 & 0 & 0.191 \\
            VGG Flower~\citep{nilsback2008automated} & 102 & 71 & 15 & 16 & 8,189 & 5,655 & 1,109 & 1,425 & 0 & 0.177 \\
            MSCOCO~\citep{lin2014microsoft} & 80 & - & 40 & 40 & 860,001 & - & 513,021 & 346,980 & 0 & 0.173 \\
            Quick Draw~\citep{jongejan2016quick} & 345 & 241 & 52 & 52 & 50,426,266 & 34,776,331 & 7,939,640 & 7,710,295 & 0 & 0.168 \\
            % \hdashline[1pt/1pt]
            CropDiseases~\citep{mohanty2016using} & 38 & - & - & 38 & 43,456 & - & - & 43,456 & 0 & 0.213 \\
            ChestX~\citep{wang2017chestx} & 8 & - & - & 8 & 25,848 & - & - & 25,848 & 0 & 0.183 \\
            EuroSAT~\citep{helber2019eurosat} & 10 & - & - & 10 & 27,000 & - & - & 27,000 & 0 & 0.173 \\
            ISIC2018~\citep{codella2019skin} & 7 & - & - & 7 & 10,015 & - & - & 10,015 & 0 & 0.186 \\
            \hdashline[1pt/1pt]
            MetaCoCo (Ours) & 100 & 64 & 16 & 20 & 175,637 & 156,666 & 5,839 & 12,268 & 155 & 0.142 \\
            \shline
        \end{tabular}
    }
\vspace{-3mm} 
\label{tab:data}
\end{table}

In Table~\ref{tab:data}, we summarize the statistics of the existing benchmarks and our proposed spurious-correlation benchmark, \textit{i.e.}, MetaCoCo. Specifically, Meta-dataset~\citep{triantafillou2019meta} and BSCD-FSL~\citep{GuoCKCSSRF20} are two commonly used cross-domain benchmarks, where Meta-dataset has \textit{10 existing datasets}, including ILSVRC-2012~\citep{deng2009imagenet}, Omniglot~\citep{lake2015human}, Aircraft~\citep{wah2011caltech}, CUB-200-2011~\citep{wah2011caltech}, Describable Textures~\citep{cimpoi2014describing}, Quick Draw~\citep{jongejan2016quick}, Fungi~\citep{schroeder2018fgvcx}, VGG Flower~\citep{nilsback2008automated}, Traffic Signs~\citep{houben2013detection} and MSCOCO~\citep{lin2014microsoft}. 
BSCD-FSL~\citep{GuoCKCSSRF20} has \textit{4 existing datasets}, including CropDiseases~\citep{mohanty2016using}, EuroSAT~\citep{helber2019eurosat}, ISIC2018~\citep{codella2019skin,tschandl2018ham10000}, and ChestX~\citep{wang2017chestx}. 
The main differences between cross-domain benchmarks and our proposed MetaCoCo benchmark are as follows: 
(1) \textbf{The cause of shifts.} The shifts in cross-domain benchmarks are caused by varying distributions between various datasets. Instead, the shifts in MetaCoCo are caused by varying both concepts and contexts. For example, for cross-domain shifts, the FSL model is trained on \textit{mini}ImageNet and tested on EuroSAT. Whereas for spurious-correlation shifts, the FSL model is trained and tested on images that have distinct associations with the contexts.
(2) \textbf{The use of contexts.} In contrast to the existing few-shot classification benchmarks, as shown in Table~\ref{tab:data}, the proposed MetaCoCo benchmark further uses context information collected from real-world scenarios to reflect the spurious-correlation shifts. 
% In fact, many data collections have a severe long-tail phenomenon in real-world applications, which results in severe coupling of context and concept information. %Therefore, in this paper, we propose a new benchmark with more context knowledge (almost 155 contexts) to simulate the spurious-correlation shifts in the real world.

\subsection{Similarity Between the Concept and Context Information}
\label{subsec:similarity-1}
\begin{figure}[tbp]
    \centering
    \subfloat[$\mathcal{M}_{ce}$ on the existing benchmarks and MetaCoCo]{
    \begin{minipage}[t]{0.55\linewidth}
        \centering
        \includegraphics[scale = 0.3]{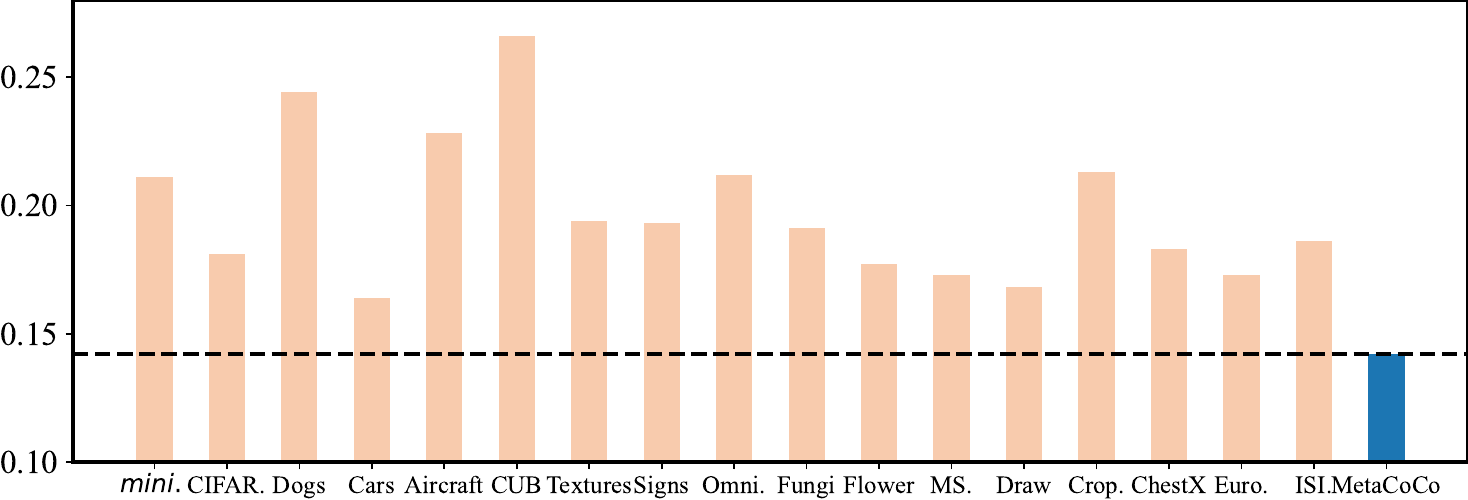}
    \end{minipage}
    }
    \subfloat[$\mathcal{M}_{ce}$ and $\mathcal{M}_{te}$ on MetaCoCo]{
    \begin{minipage}[t]{0.35\linewidth}
        \centering
        \includegraphics[scale = 0.244]{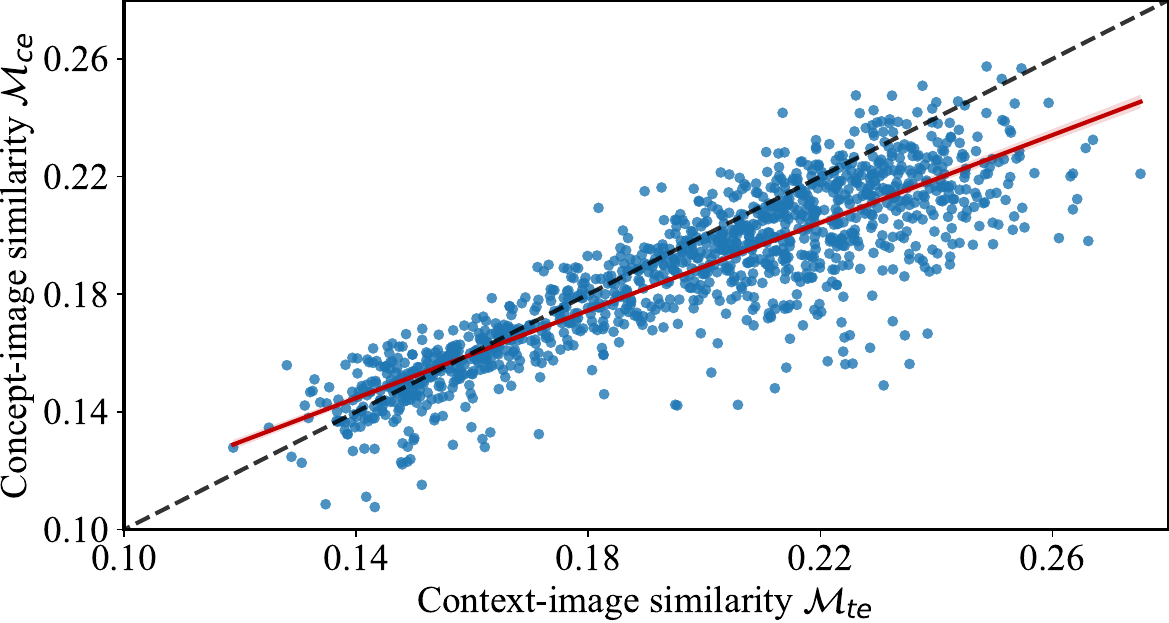}
    \end{minipage}
    }
\centering
\vspace{-2mm}
\caption{(a) The sample-averaged similarity $\mathcal{M}_{ce}$ between concepts and images on the existing FSC benchmarks and the proposed MetaCoCo, where MetaCoCo has significantly lower similarity between contexts and images. (b) The context-image similarities $\mathcal{M}_{te}$ (horizontal axis) versus the concept-image similarities $\mathcal{M}_{ce}$ (vertical axis) of the sample points in the MetaCoCo.}
\label{fig:2}
\vspace{-3mm}
\end{figure}

For images containing both conceptual and contextual information, a greater similarity between image and context implies that the benchmark has more spurious-correlation shifts. To intuitively show that MetaCoCo has considerably more spurious-correlation shifts than the existing FSC benchmarks including cross-domain-shift benchmarks, we introduce a novel metric that uses CLIP~\citep{radford:clip} as a pre-trained vision-language model. By calculating the cosine distance of text and image features extracted by pre-trained text and image encoder from CLIP, the similarity $\mathcal{M}_{ce}$ between \textcolor{red}{conceptual language information} and {image visual knowledge}, and the similarity $\mathcal{M}_{te}$ between \textcolor{blue}{contextual language expression} and {image visual knowledge} are calculated as follows:
%Each image has concept and context information and if there is a large similarity between the image and context, it means that the benchmark has a significantly spurious-correlation shift. 
%Therefore, to evaluate that our MetaCoCo has more spurious-correlation shifts than existing few-shot classification (FSC) benchmarks including cross-domain-shift benchmarks, we introduce a novel metric method. Motivated by vision-language model like CLIP~\citep{radford:clip}, it calculates the cosine distance of text and image features extracted by pre-trained text and image encoder from CLIP. 
%The similarity between conceptual language information and image visual knowledge $\mathcal{M}_{ce}$ or contextual language expression and image visual information $\mathcal{M}_{te}$ is calculated as follows: 
\begin{equation}
   \mathcal{M}_{ce} = d(z_{x}, \textcolor{red}{z_{t}^{ce}}), \quad 
   \mathcal{M}_{te} = d(z_{x}, \textcolor{blue}{z_{t}^{te}}),
\end{equation}
where $d(\cdot, \cdot)$ is the cosine distance measurement, $z_{x}$ is the {image} features extracted by pre-trained image encoder by CLIP, and $\textcolor{red}{z_{t}^{ce}}$ and $\textcolor{blue}{z_{t}^{te}}$ represent the text features of \textcolor{red}{concept} and \textcolor{blue}{context} extracted by pre-trained text encoder by CLIP, respectively. %The similar results are shown in Figure~\ref{fig:similarity-2}.
% Figure~\ref{fig:similarity-1} denotes the similarity between the text information of concept and the image knowledge of image on each sample compared with MetaCoCo and existing 16 datasets. It shows that the similarity between the concept and image from MetaCoCo is weak. In other words, the image and label matching of each sample of MetaCoCo are poor.
Figure~\ref{fig:2}(a) shows the sample-averaged similarity $\mathcal{M}_{ce}$ between concepts\footnote{Since the existing FSC benchmarks lack context information as shown in Table \ref{tab:data}, we are not able to compute their sample-averaged similarity $\mathcal{M}_{te}$ between contexts and images.} and images on the existing FSC benchmarks as well as the proposed MetaCoCo. It can be seen that MetaCoCo has significantly lower similarity between concepts and images. This is because the added context information in the image introduces spurious-correlations with the concepts, \textit{e.g.}, ``grass" and ``dog", thus weakening the direct correlation between the images and the concepts or labels, and presenting a more challenging evaluating benchmark for the FSC. Figure~\ref{fig:2}(b) further shows the context-image similarities $\mathcal{M}_{te}$ (horizontal axis) versus the concept-image similarities $\mathcal{M}_{ce}$ (vertical axis) of the sample points in the MetaCoCo. We find that the overall context-image similarities are slightly higher than the concept-image similarities, suggesting that spurious-correlation shifts are substantial in the proposed benchmark.

% \begin{figure}[tbp]
%   \centering
%   \vspace{-1mm}
%   \includegraphics[height=0.23\textwidth]{figures/metacoco_sim_all.pdf}
%   \includegraphics[height=0.23\textwidth]{figures/metacoco_sim_train.pdf}
%   \includegraphics[height=0.23\textwidth]{figures/metacoco_sim_val.pdf}
%   \includegraphics[height=0.23\textwidth]{figures/metacoco_sim_test.pdf}
%   \vspace{-2mm}
%   \caption{Similarity of each concept-image pair (vertical axis) or context-image pair (horizontal axis) on MetaCoCo. 
%   These figures represent total, training, validation and testing from left to right.}
%   \label{fig:similarity-2}
%   \vspace{-5mm}
% \end{figure}

\subsection{Evaluation Strategies}
\label{subsec:eval}

Before presenting the datasets, we first discuss the evaluation strategies in MetaCoCo, including:

(1) \textbf{Fine-tuning based methods.} Fine-tuning based methods follow the transfer learning procedure, including two phases: pre-training with base classes and test-tuning with novel classes. In the pre-training with base classes phase, the base classes $\mathcal{D}_{b}$ is used to train a $\mathcal{C}_{base}$-class classifier as below:
\begin{equation} \mathnormal{\Gamma}=\underset{\mathcal{\theta,\phi}}{\arg\min}\sum_{i=1}^{T}\mathcal{L}_\text{CE}\Big(c_{\phi}(f_{\theta}(x_{i})),y_{i}\Big)\,,
\label{eq:fine-tuning}
\end{equation}
where $T$ is the sample number of $\mathcal{D}_{b}$, and $\mathcal{L}_\text{CE}(\cdot, \cdot)$ is the cross-entropy loss. In the test-tuning with novel classes phase, each episode $\mathcal{T}_{e}=\langle\mathcal{S}_{e}, \mathcal{Q}_{e}\rangle$ is sampled from novel classes $\mathcal{D}_{n}$ and a new $\mathcal{C}_{e}$-class classifier is re-learned based on a few labeled images $\mathcal{S}_{e}$ and tested on $\mathcal{Q}_{e}$. 
% Basically, the pre-trained feature encoder $f_{\theta}$ is fixed to avoid over-fitting. 

(2) \textbf{Metric-based methods.}
Metric-based methods directly compare the similarities (or distance) between query images and support classes, \textit{i.e.}, learning to compare, through the episodic training mechanism. 
Taking Prototypical Network (ProtoNet)~\citep{snell2017prototypical} as an example, it takes the mean vector of each support class as its corresponding prototype representation, and then compares the relationships between query images and prototypes. The prototype $p_{n}$ of each class in the support set $\mathcal{S}_{e}$ can be formulated as $p_{n} = \frac{1}{K}\sum_{(x_i, y_i)\in\mathcal{S}_e}f_{\theta}(x_{i})\cdot\mathbb{I}(y_i=n)$, where  $\mathbb{I}(\cdot)$ is the indicator function, then the the metric loss on $\mathcal{Q}_e$ can be computed as:
\begin{equation}
        \mathcal{L}(\theta)=-\frac{1}{I_{q}}\sum_{(x_i, y_i)\in\mathcal{Q}_e}\log P(y_i|\mathcal{Q}_{e}),\quad   \text{where} \ P(y_i|\mathcal{Q}_{e}) = \frac{\exp(-D(f_{\theta}(x_{i}),p_{y_{i}}))}{\sum_{n=1}^N\exp(-D(f_{\theta}(x_{i}), p_{n}))},
\end{equation}
and $D(\cdot, \cdot)$ denotes a distance measurement, \textit{e.g.}, the squared euclidean distance in the ProtoNet. 

(3) \textbf{Meta-based methods.}
Meta-based methods aim to make the trained model able to quickly adapt to unseen novel tasks by a few gradient steps in the testing phase. Specifically, the learning paradigm of meta-based methods has two levels, \textit{i.e.}, inner-level and outer-level, to update the base and meta learner, respectively. Model-agnostic meta-learning (MAML)~\citep{FinnAL17} is one representative method, whose core idea is to train a model's initial parameters by using the two levels. Specifically, the base learner is optimized on the support set $\mathcal{S}_{e}$ that
\begin{equation}
    \begin{aligned}
        \{\theta, \phi\} &\leftarrow \{\theta, \phi\} - \eta_{out} \nabla_{\{\theta, \phi\}} \mathcal{L}_{ce}(c_{\phi^{\prime}}(f_{\theta^{\prime}}(x_{i}), y_{i})), \\
        \text{where} \quad \{\theta^{\prime}, \phi^{\prime}\} &= \{\theta, \phi\} - \eta_{in} \nabla_{\{\theta, \phi\}} \mathcal{L}_{ce}(c_{\phi}(f_{\theta}(x_{i}), y_{i})), 
    \end{aligned}
\end{equation}
and the $\eta_{in}$ and $\eta_{out}$ are the learning rates of the inner level and the outer level, respectively.

\section{MetaCoCo: A New Few-Shot Classification Benchmark with Spurious Correlation}
\label{sec:data}
MetaCoCo aims to present an environment for evaluating the fine control of spurious-correlation shifts in the FSC problems. Specifically, our approach consists of (1) dataset generating, and (2) episode sampling, whose operational procedures are detailed below.
% MetaCoCo aims to offer an environment for measuring progress in spurious-correlation few-shot classification tasks. Our approach is twofold: (1) changing the data and (2) changing the formulation of the task (\textit{i.e.}, how episodes are generated). The following sections describe these modifications.

\textbf{Dataset generating.}
%Compared with previous benchmarks, MetaCoCo's data consists of concept and context information, and each concept has a strong correlation with the context, which leads to distribution shifts between training and testing data. 
Compared with the existing benchmarks, the samples in MetaCoCo consist of both conceptual and contextual information, and many of these images exhibit a strong correlation with the context, which increases the impact of spurious-correlation shifts between the training data and the testing data on the prediction performance. Specifically, we first select 100 categories of common objects following DomainNet~\citep{peng2019moment}. These categories include 155 contexts, which are collected from the adjectives or nouns appeared more frequently with these categories from WordNet~\citep{miller1995wordnet}. Then the images are collected by searching a category name combined with a context name (\textit{e.g.}, ``dog on grass") in various image search engines. One of the main challenges is that the downloaded data contains a large portion of outliers. To clean the dataset, we manually filter out the outliers, which takes around 2,500 hours in total. To control the annotation quality, we assign two annotators to each image and only take the images agreed by both annotators. After the filtering process, we kept 17.6k images from the 1.0 million images crawled from the web. The dataset has an average of around 1,000 images per category  
(see Appendix~\ref{app-b} for more details).
% (see Appendix B for more details).

\textbf{Episode sampling.}
MetaCoCo has 100 categories, and the number of matching contexts for each category is inconsistent, resulting in an inconsistent number of samples for each category. We sort the samples from most to least. The first 64 categories with the largest number of samples are used as training data, then 20 categories are selected as testing data, and the last 16 categories are used as validation data.
FSC adopts an episodic paradigm to train and test the model. Each $N$-way $K$-shot episode $\mathcal{T}_{e}$ has a support set $\mathcal{S}_{e}$ and a query set $\mathcal{Q}_{e}$, where $\mathcal{S}_{e}$ and $\mathcal{Q}_{e}$ share the same categories but different images.
Therefore, we have two sample episodic strategies: independent and identically distributed (IID) episode, \textit{i.e.}, the support and query images with the \emph{same} contexts, and out-of-distribution (OOD) episode, \textit{i.e.}, the support and query images with the \emph{different} contexts.
\section{Experiments}
\label{sec:exp}

In this section, we evaluate the spurious-correlation performance of the state-of-the-art methods optimized with different learning strategies. These experiments further demonstrate that SC-FSC is still a major challenge.
(see Appendix~\ref{app-c} and~\ref{app-d} for more experimental details and results).

\subsection{Experimental Setup}
\label{subsec:set}

\begin{table}[tbp]  
    \centering
    \caption{Experiments in state-of-the-art few-shot classification and self-supervised learning methods. ``rot." and ``jig." mean using the Rotation and Jigsaw self-supervised pretext tasks, respectively.}
    \vspace{-3mm}
    % \scalebox{0.86}{
    \resizebox{1.0\textwidth}{!}{
        \begin{tabular}{llccccc|cc}
            \shline
            Method & Conference & Backbone & Type & GL & LL & TT & $1$-shot & $5$-shot \\ 
            \hline
            Baseline~\citep{ChenLKWH19} & ICLR 2019     
            & ResNet12 & Fine-tuning & \cmark & & \cmark & 46.78 & 60.78 \\
            Baseline++~\citep{ChenLKWH19} & ICLR 2019 & ResNet12 & Fine-tuning  & \cmark & & \cmark & 46.95 & 58.50  \\
            RFS-simple~\citep{tian2020rethinking} & ECCV 2020 & ResNet12 & Fine-tuning & \cmark & & \cmark & 47.02   &  56.71       \\
            Neg-Cosine~\citep{liu2020negative} & ECCV 2020 & ResNet12 & Fine-tuning & \cmark & & \cmark & 50.78  &  62.34        \\
            SKD-GEN0~\citep{rajasegaran2020self} & BMVC 2021 & ResNet12 & Fine-tuning  & \cmark & & \cmark &  51.34  &  63.21    \\
            FRN~\citep{wertheimer2021few} & CVPR 2021 & ResNet12 & Fine-tuning  & \cmark & & \cmark & 50.23 & 60.56  \\
            Yang et al~\citep{yang2022few} & ECCV 2022 & ResNet12 & Fine-tuning  & \cmark & & \cmark & 58.01 & 69.32 \\
            LP-FT-FB~\citep{wang2022revisit} & ICLR 2023 & ResNet12 & Fine-tuning & \cmark & \cmark & \cmark & 56.21 & 70.21 \\ 
            \hdashline[1pt/1pt]
            MAML~\citep{FinnAL17} & ICML 2017          
            & ResNet12 & Meta &  & \cmark & \cmark & 45.01  & 54.21   \\
            Versa~\citep{gordon2018versa}  & NeurIPS 2018 & ResNet12  & Meta  &  & \cmark & \cmark & 39.64 & 53.06  \\
            R2D2~\citep{bertinetto2018meta} & ICLR 2019
            & ResNet12  & Meta &  & \cmark & \cmark & 45.25 & 60.14         \\
            MTL~\citep{sun2019meta} & CVPR 2019 & ResNet12 & Meta & \cmark & \cmark & \cmark &  44.23 &  58.04    \\
            ANIL~\citep{RaghuRBV20} & ICLR 2020    
            & ResNet12  & Meta  &  & \cmark & \cmark   & 36.58 & 50.54             \\
            BOIL~\citep{oh2020boil} & ICLR 2021 & ResNet12 & Meta   &  & \cmark & \cmark    &     44.09 &  55.61     \\
            CDKT+ML~\citep{ke2023revisiting} & NeurIPS 2023 & ResNet18 & Meta & & \cmark & \cmark & 44.86 & 61.42 \\
            CDKT+PL~\citep{ke2023revisiting} & NeurIPS 2023 & ResNet18 & Meta & & \cmark & \cmark & 43.21 & 59.87 \\
            \hdashline[1pt/1pt]
            % ProtoNet~\citep{snell2017prototypical} & NeurIPS 2017 & ResNet12   & Metric  & \cmark & &  & 42.69 & 59.50      \\
            % RelationNet~\citep{sung2017learning} & CVPR 2018 & ResNet12 & Metric  & & \cmark & & 45.32 & 57.73       \\
            CovaMNet~\citep{li2019distribution} & AAAI 2019 & ResNet12  & Metric  & \cmark & &  & 47.81 & 58.43   \\
            DN4~\citep{li2019revisiting}  & CVPR 2019   & ResNet12  & Metric   & \cmark & &   
            & 45.04 & 57.68      \\
            CAN~\citep{hou2019cross} & NeurIPS 2019  & ResNet12  & Metric   & \cmark & \cmark & & 48.93 & 62.36              \\
            % RENet~\citep{kang2021relational} & CVPR 2018  & ResNet12  & Metric   & \cmark & \cmark &  & 50.21   &  64.23   \\
            DeepBDC~\citep{xie2022joint} & CVPR 2022 & ResNet12 & Metric & & \cmark & \cmark & 46.78 & 62.54 \\
            FGFL~\citep{cheng2023frequency} & ICCV 2023 & ResNet12 & Metric &  & \cmark & \cmark & 46.78 & 64.32 \\
            PUTM~\citep{tian2023prototypes} & ICCV 2023 & ResNet18 & Metric & & \cmark & \cmark & 60.23 & 72.36 \\
            TSA+DETA~\citep{zhang2023deta} & ICCV 2023 & ResNet18 & Metric & & \cmark & \cmark & 51.42 & 61.58 \\ 
            \hline\hline
            MoCo~\citep{he2020momentum} & CVPR 2020 & ResNet50 & Self-supervised learning & 
            & \cmark & \cmark & 56.90 & 70.65 \\
            SimCLR~\citep{chen2020simple} & ICML 2020 & ResNet50 & Self-supervised learning & 
            & \cmark & \cmark & 58.12 & 71.21 \\
            \hdashline[1pt/1pt]
            ProtoNet~\citep{snell2017prototypical} & NeurIPS 2017 & ResNet18 & Metric & & \cmark & & 43.14 & 57.84 \\
            \quad + rot. + SSFSL~\citep{su2020does} & ECCV 2020 & ResNet18 & Self-supervised learning & & \cmark & & 40.65 & 54.31 \\
            \quad + rot. + HTS~\citep{zhang2022tree} & ECCV 2022 & ResNet18 & Self-supervised learning & & \cmark & & 42.06 & 55.13 \\
            \quad + jig. + SSFSL~\citep{su2020does} & ECCV 2020 & ResNet18 & Self-supervised learning & \cmark & & & 45.43 & 58.91 \\
            \quad + rot. + jig. + SSFSL~\citep{su2020does} & ECCV 2020 & ResNet18 & Self-supervised learning & & \cmark & & 44.46 & 59.01 \\
            \hdashline[1pt/1pt]
            ProtoNet~\citep{snell2017prototypical} &  NeurIPS 2017 & ResNet12 & Metric & & \cmark & & 42.69 & 59.50   \\
            \quad + rot. + SLA~\citep{lee2020self} & ICML 2020 & ResNet12 & Self-supervised learning & & \cmark & & 40.29 & 58.09  \\
            \quad + rot. + HTS~\citep{zhang2022tree} & ECCV 2022 & ResNet12 & Self-supervised learning & & \cmark & \cmark & 43.19 & 60.50  \\
            \hdashline[1pt/1pt]
            ProtoNet~\citep{snell2017prototypical} & NeurIPS 2017 & WRN-28-10  & Metric  & & \cmark & & 43.67 & 60.78 \\
            \quad + rot. + BF3S~\citep{gidaris2019boosting} & ICCV 2019 & WRN-28-10 & Self-supervised learning &  & \cmark &  & 43.78 & 57.64  \\
            \quad + rot. + HTS~\citep{zhang2022tree} & ECCV 2022 & WRN-28-10 & Self-supervised learning  & & \cmark &  & 45.31 & 62.31  \\
            \shline
    \end{tabular}
} 
\label{tab:fsc_acc}
\vspace{-4mm}
\end{table}  

\noindent \textbf{Few-shot classification methods.}
We evaluate the performance with a large number of algorithms that span different learning strategies, including: (1) Five \textit{fine-tuning based methods}: Baseline~\citep{ChenLKWH19}, Baseline++~\citep{ChenLKWH19}, RFS-simple~\citep{tian2020rethinking}, Neg-Cosine~\citep{liu2020negative} and SKD-GEN0~\citep{rajasegaran2020self}.  
(2) Six \textit{metric-based methods}: ProtoNet~\citep{snell2017prototypical}, RelationNet~\citep{sung2017learning}, CovaMNet~\citep{li2019distribution}, DN4~\citep{li2019revisiting}, CAN~\citep{hou2019cross} and RENet~\citep{kang2021relational}. 
(3) Six \textit{meta-based methods}: MAML~\citep{FinnAL17}, Versa~\citep{gordon2018versa}, R2D2~\citep{bertinetto2018meta}, MTL~\citep{sun2019meta}, ANIL~\citep{RaghuRBV20} and BOIL~\citep{oh2020boil}.        
(4) Six \textit{self-supervised learning methods}: MoCo~\citep{he2020momentum}, SimCLR~\citep{chen2020simple}, SSFSL~\citep{su2020does}, HTS~\citep{zhang2022tree}, SLA~\citep{lee2020self} and BF3S~\citep{gidaris2019boosting}.
(5) Seven \textit{cross-domain methods}: Linear~\citep{YueZS020}, Cosine~\citep{YueZS020}, \textit{k}-NN~\citep{YueZS020}, ATA~\citep{WangD21}, FT~\citep{TsengLH020}, LRP~\citep{sun2021explanation} and IFSL~\citep{YueZS020}.
% (6) four \textit{Parameter-efficient methods}: CLIP~\citep{radford:clip}, CoOp~\citep{zhou:coop}, CoCoOp~\citep{zhou:cocoop} and Linear Probing~\citep{radford:clip}.

\noindent \textbf{Backbone architectures.}
Following prior literatures~\citep{li2021libfewshot}, all fine-tuning based methods, metric-based methods and meta-based methods adopt three different embedding backbones from shallow to deep, \textit{i.e.}, Conv64F, ResNet12 and ResNet18. For other learning strategy methods, we adopt different feature backbones based on the corresponding original papers, \textit{e.g.}, ResNet10 for cross-domain few-shot classification methods, WRN-28-10 for self-supervised learning methods. 
% or ResNet50 for parameter-efficient methods.

\noindent \textbf{Evaluation protocols.}
Following the prior work~\citep{li2021libfewshot}, in this paper, we control the evaluation for all methods 
% except the parameter-efficient methods
, evaluate them on 600 sampled tasks and repeat this process five times, \textit{i.e.}, a total of 3,000 tasks. The top-1 mean accuracy will be reported. All images are resized into 84 $\times$ 84 by using the single center crop~\citep{li2019distribution}. 
Three common tricks are used: (1) \textit{Global-label (GL)} indicates that the global labels of the training set are used for pre-training during the training phase. (2) \textit{Local-label (LL)} means that only the specific local labels are used in the episodic training phase. (3) \textit{Test-tune (TT)} means test-tuning of using the support set at the testing stage.
% For parameter-efficient methods, the experimental setting follows CLIP.

\subsection{Main results} 
In this section, we conduct extensive experiments on various methods with six learning strategies.

\begin{table}[tbp]  
    \centering
    \caption{Experiments of cross-domain and spurious-correlation few-shot classification methods. }
    \vspace{-3mm}
    % \scalebox{0.92}{
    \resizebox{0.95\textwidth}{!}{
        \begin{tabular}{llcccc|cc}
            \shline
            Method & Conference & Type & GL & LL & TT & $5$-way $1$-shot & $5$-way $5$-shot  \\ 
            \hline
            RelationNet~\citep{sung2017learning} & CVPR 2018 & Metric & & \cmark &  & 45.32 \scriptsize$\pm$ 0.48 & 57.73 \scriptsize$\pm$ 0.45\\
            \quad +ATA~\citep{WangD21} & IJCAI 2021       
            & CD-FSC & & \cmark & & 43.24 \scriptsize$\pm$ 0.47 & 56.94 \scriptsize$\pm$ 0.47 \\
            \quad +FT~\citep{TsengLH020} & ICLR 2020          
            & CD-FSC & & \cmark &  & 45.37 \scriptsize$\pm$ 0.50 & 58.74 \scriptsize$\pm$ 0.48 \\
            \hdashline[1pt/1pt]
            GNN~\citep{satorras2018few} & ICLR 2018        
            & Metric & & \cmark &  & 48.14 \scriptsize$\pm$ 0.55 & 61.94 \scriptsize$\pm$ 0.56   \\
            \quad +ATA~\citep{WangD21} & IJCAI 2021        
            & CD-FSC &  & \cmark & & 46.78 \scriptsize$\pm$ 0.55 &  61.78 \scriptsize$\pm$ 0.52  \\
            \quad + FT~\citep{TsengLH020} & ICLR 2020          
            & CD-FSC &  & \cmark &  &  47.30 \scriptsize$\pm$ 0.56 & 65.90 \scriptsize$\pm$ 0.56  \\
            % \quad +LRP~\citep{sun2021explanation} & ICPR 2021 & CD-FSC &  & \cmark &  & 48.62 \scriptsize$\pm$ 0.54 & 62.76 \scriptsize$\pm$ 0.54  \\
            \hdashline[1pt/1pt]
            TPN~\citep{liu2018learning} & ICLR 2019 & Metric & & \cmark &  & 49.65 \scriptsize$\pm$ 0.51 & 60.62 \scriptsize$\pm$ 0.47\\
            \quad +ATA~\citep{WangD21} & IJCAI 2021 & CD-FSC & & \cmark & & 47.15 \scriptsize$\pm$ 0.53 &  60.33 \scriptsize$\pm$ 0.31  \\
            \quad +FT~\citep{TsengLH020} & ICLR 2020         
            & CD-FSC &  & \cmark &  & 45.62 \scriptsize$\pm$ 0.51 & 55.78 \scriptsize$\pm$ 0.52  \\
            \hdashline[1pt/1pt]
            Linear~\citep{YueZS020} & NeurIPS 2020 & Fine-tuning & &   & & 43.31 \scriptsize$\pm$ 0.40 & 57.87 \scriptsize$\pm$ 0.41 \\
            Cosine~\citep{YueZS020} & NeurIPS 2020 & Fine-tuning  & \cmark &  & \cmark & 42.81 \scriptsize$\pm$ 0.42 & 56.33 \scriptsize$\pm$ 0.41\\
            \textit{k}-NN~\citep{YueZS020} & NeurIPS 2020 & Fine-tuning & \cmark &  & \cmark & 42.22 \scriptsize$\pm$ 0.42 & 57.93 \scriptsize$\pm$ 0.42  \\
            \hdashline[1pt/1pt]
            MAML~\citep{FinnAL17} & ICML 2017 & Meta  
            & & \cmark & \cmark & 44.09 \scriptsize$\pm$ 0.52 &  53.98 \scriptsize$\pm$ 0.48  \\
            \quad +IFSL~\citep{YueZS020} & NeurIPS 2020 & SC-FSC & & \cmark & \cmark & 43.42 \scriptsize$\pm$ 0.51 & 55.00 \scriptsize$\pm$ 0.48  \\
            MTL~\citep{sun2019meta} & CVPR 2019      
            & Meta  & \cmark & \cmark & \cmark & 43.80 \scriptsize$\pm$ 0.48 & 57.18 \scriptsize$\pm$ 0.48 \\
            \quad +IFSL~\citep{YueZS020} & NeurIPS 2020          
            & SC-FSC & \cmark & \cmark & \cmark & 43.42 \scriptsize$\pm$ 0.48 & 56.90 \scriptsize$\pm$ 0.48  \\
            MatchingNet~\citep{VinyalsBLKW16} & NeurIPS 2016 
            & Metric & & \cmark &  & 43.72 \scriptsize$\pm$ 0.49 & 56.12 \scriptsize$\pm$ 0.49 \\
            \quad +IFSL~\citep{YueZS020}  & NeurIPS 2020         
            & SC-FSC  & & \cmark &  & 44.11 \scriptsize$\pm$ 0.49 & 55.86 \scriptsize$\pm$ 0.49   \\
            SIB~\citep{hu2020empirical} & ICLR 2020         
            & Meta  & & \cmark & \cmark & 48.43 \scriptsize$\pm$ 0.57 & 58.53 \scriptsize$\pm$ 0.51   \\
            \quad +IFSL~\citep{YueZS020} & NeurIPS 2020          
            & SC-FSC  & & \cmark & \cmark & 47.97 \scriptsize$\pm$ 0.54 & 58.41 \scriptsize$\pm$ 0.50   \\
            \shline
        \end{tabular}
    } 
\label{tab:cdfsc}
\vspace{-1mm}
\end{table}

\begin{figure}[t]
  \centering
  \vspace{-0.3mm}
  \includegraphics[height=0.19\textwidth]{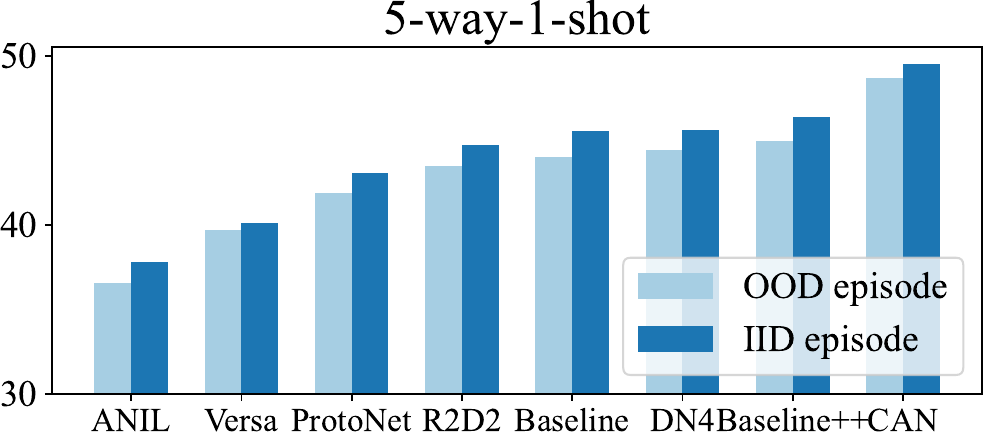}
  \includegraphics[height=0.19\textwidth]{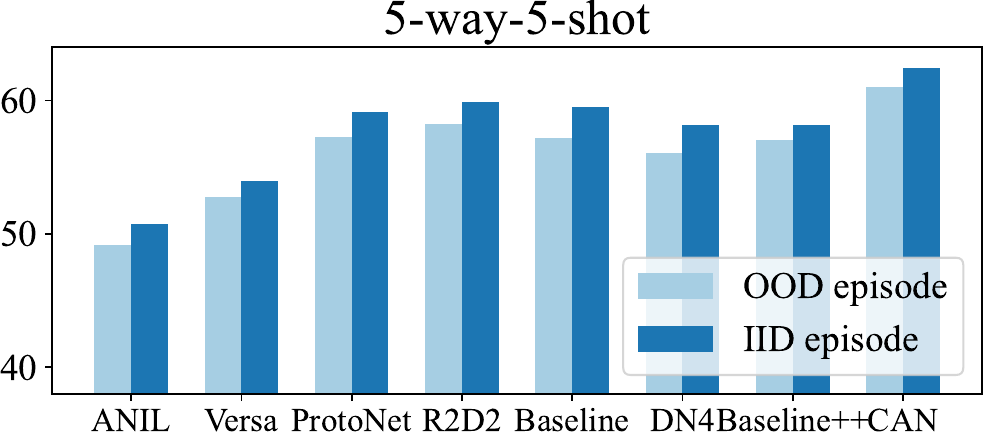}
  \vspace{-2mm}
  \caption{Experiments of the test-tuning phase with different sampling episodes, \textit{i.e.}, IID and OOD.}
  \label{fig:iid-ood-episode}
  \vspace{-4mm}
\end{figure}

\textbf{Experiments in fine-tuning, metric- and meta-based methods and self-supervised methods.}
We evaluate the performance of 17 competing few-shot methods and six self-supervised methods in our 
MetaCoCo. The results of the 5-way 1- or 5-shot setting are shown in Table~\ref{tab:fsc_acc}. 
From Table~\ref{tab:fsc_acc}, we have the following findings: (1) We find that the performance of all methods decreases compared with existing FSC benchmarks~\citep{li2021libfewshot}, which demonstrates that these methods are insufficient in solving the spurious-correlation-shift problem. (2) Previous works introduced self-supervised learning to improve the generalization of FSC models, but experiments have shown that this is not suitable for the SC-FSC problem. In some cases, using self-supervised learning even damages the performance, \textit{i.e.}, ProtoNet has 43.14\% in 1-shot, but the accuracy by using rotation is 40.65\%. 
% We include more detailed versions of the table displaying 95\% confidence intervals in the Appendix~\ref{app-d}. 

\textbf{Experiments in CD-FSC and SC-FSC methods.}
Table~\ref{tab:cdfsc} displays the accuracy of seven CD-FSC methods. These methods have a significant performance on solving the cross-domain-shift problem on the {Meta-dataset}~\citep{triantafillou2019meta} and BSCD-FSL~\citep{GuoCKCSSRF20}. However, in MetaCoCo, the advantages of these methods disappear, resulting in weaker performance, even worse than non-cross-domain FSC methods. It is worth noting that the main motivation of IFSL~\citep{YueZS020} is to use the idea of causality to solve the impact of spurious correlation between contextual information and images on the model training phase. However, we observe a substantial decrease of the performance on the real-world spurious-correlation benchmark, \textit{i.e.}, MetaCoCo.

\begin{figure}[tbp]
  \centering
  \vspace{-1mm}
  \includegraphics[height=0.20\textwidth]{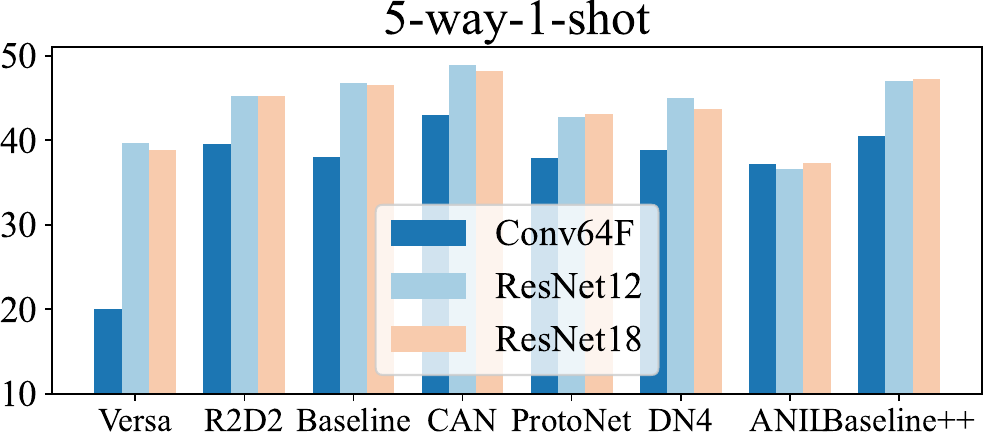}
  \includegraphics[height=0.20\textwidth]{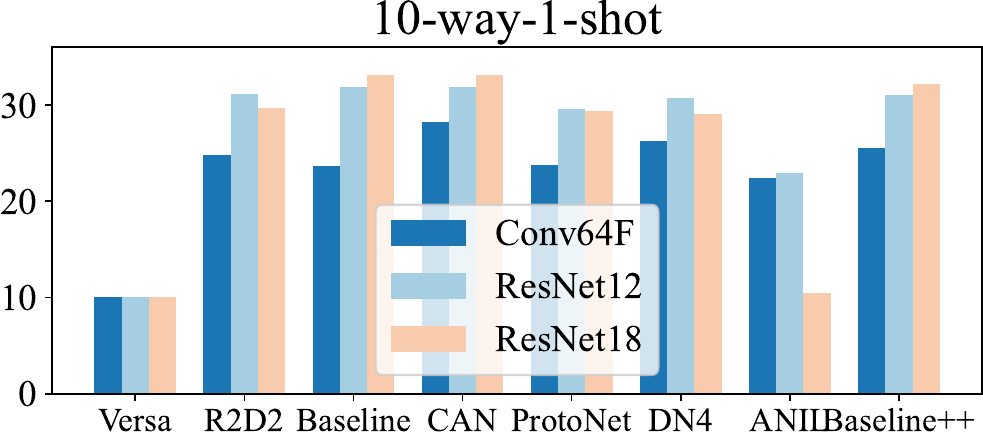}
  \vspace{-2mm}
  \caption{Experiments of different backbone architectures under 5-way and 10-way 1-shot settings.}
  \label{fig:backbones}
  \vspace{-2.5mm}
\end{figure}
\begin{figure}[tbp]
  \centering
  \vspace{-1mm}
  \includegraphics[height=0.18\textwidth]{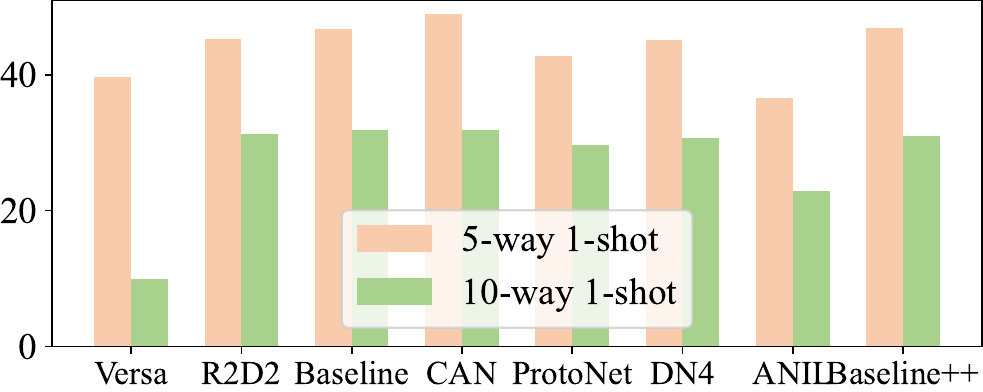}
  \includegraphics[height=0.18\textwidth]{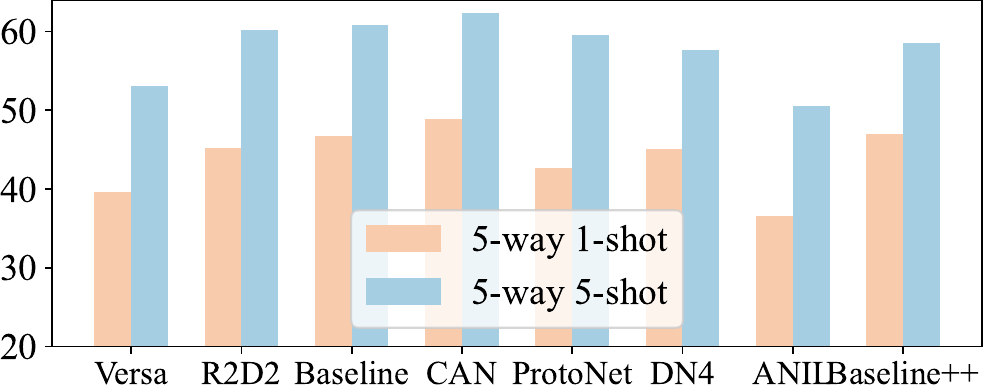}
  \vspace{-2mm}
  \caption{Experimental results of different ways (left) and shots (right) on testing performance.}
  \label{fig:way-shot}
  \vspace{-3mm}
\end{figure}

% \textbf{Experiments in parameter-efficient methods.}
To this end, according to these experimental results, we observe that most methods are insufficient to solve the spurious-correlation-shift FSC problem.  
We hope the proposed MetaCoCo can facilitate future research on the important and real-world problem for few-shot classification.

\subsection{In-depth study}
To further analyze the influence of spurious shifts in MetaCoCo, we conduct in-depth experiments.

\textbf{Effect of the IID and OOD episodes.}
Figure~\ref{fig:iid-ood-episode} shows the results of FSC methods under 5-way 1- and 5-shot settings. The IID and OOD episodes represent the same and different contexts of the support and query sets during the test-tuning phase, respectively (see Section~\ref{sec:data}). These results clearly denote that the learning process of the IID episode is better than the optimization process of the OOD episode.
This further demonstrates that the model tends to utilize contextual information during the learning process. Once images do not match the contexts, the performance will deteriorate.

\textbf{Effect of different backbone architectures.}
In \cite{ChenLKWH19}, they change the depth of the feature backbone to reduce intra-class variation for all methods. Following this paper, we start from Conv64F and gradually increase the backbone to ResNet12 and 18. The experiments under 5-way and 10-way 1-shot settings are shown in Figure~\ref{fig:backbones}. It is arguably a common sense that the stronger backbone is used, the performance is best. However, we surprisingly find that this may not be always in the SC-FSC problem. Figure~\ref{fig:backbones} shows the performance degradation in some settings.

\textbf{Ways and shots analysis.}
We further study the performance of ``ways" (Figure~\ref{fig:way-shot} left) and ``shots" (Figure~\ref{fig:way-shot} right). 
As expected, we found that the difficulty increases as the way increases, and performance degrades. More examples per class, on the other hand, indeed make it easier to correctly classify that class. 
Interestingly, Versa presents a poor performance with increasing the way but it improves at a high rate when the shot increases, which further represents that the contextual effects become larger when the task becomes difficult. CAN has the best accuracy under all settings because it uses a transduction strategy to introduce query samples in the training phase, which destroys the strong spurious correlations between contexts and images. 
\section{Conclusion}
\label{sec:con}
In this paper, we present Meta Concept Context (MetaCoCo), a large-scale, diverse and realistic environment benchmark for spurious-correlation few-shot classification. We believe that our exploration of various modes on MetaCoCo has uncovered interesting directions for future works: it remains unclear what is the best learning strategy for avoiding the effect of spurious-correlation contexts and the most appropriate episodic sample.
% (\textit{e.g.}, using the pre-trained vision-language model). 
Current models even including these cross-domain FSC models don't work when trained on mismatching contexts. Current models are also not robust to the amount of data in testing episodes, each excelling in a different part of the spectrum. We believe that addressing these shortcomings constitutes an important research goal moving forward.

% \subsubsection*{Author Contributions}
% If you'd like to, you may include a section for author contributions as is done
% in many journals. This is optional and at the discretion of the authors.

\subsubsection*{Acknowledgments}
This work was supported in part by National Natural Science Foundation of China (No. U20A20387, 62376243, 62037001, 623B2002), the StarryNight Science Fund of Zhejiang University Shanghai Institute for Advanced Study (SN-ZJU-SIAS-0010) and Project by Shanghai AI Laboratory (P22KS00111). All opinions in this paper are those of the authors and donot necessarily reflect the views of the funding agencies.

% \clearpage
\bibliography{iclr2024_conference}

\begin{thebibliography}{107}
\providecommand{\natexlab}[1]{#1}
\providecommand{\url}[1]{\texttt{#1}}
\expandafter\ifx\csname urlstyle\endcsname\relax
  \providecommand{\doi}[1]{doi: #1}\else
  \providecommand{\doi}{doi: \begingroup \urlstyle{rm}\Url}\fi

\bibitem[Ahmed et~al.(2020)Ahmed, Bengio, van Seijen, and
  Courville]{ahmed2020systematic}
Faruk Ahmed, Yoshua Bengio, Harm van Seijen, and Aaron Courville.
\newblock Systematic generalisation with group invariant predictions.
\newblock In \emph{International Conference on Learning Representations,
  {ICLR}}, 2020.

\bibitem[Arjovsky et~al.(2019)Arjovsky, Bottou, Gulrajani, and
  Lopez-Paz]{arjovsky2019}
Martin Arjovsky, L{\'e}on Bottou, Ishaan Gulrajani, and David Lopez-Paz.
\newblock Invariant risk minimization.
\newblock \emph{arXiv preprint arXiv:1907.02893}, 2019.

\bibitem[Bae et~al.(2021)Bae, Choi, and Lee]{bae2021meta}
Jun-Hyun Bae, Inchul Choi, and Minho Lee.
\newblock Meta-learned invariant risk minimization.
\newblock \emph{arXiv preprint arXiv:2103.12947}, 2021.

\bibitem[Bai et~al.(2024)Bai, Zhang, Zhou, Huang, Luan, Wang, and
  Chen]{bai2024prompt}
Shuanghao Bai, Min Zhang, Wanqi Zhou, Siteng Huang, Zhirong Luan, Donglin Wang,
  and Badong Chen.
\newblock Prompt-based distribution alignment for unsupervised domain
  adaptation.
\newblock In \emph{Proceedings of the AAAI conference on artificial
  intelligence, {AAAI}}, 2024.

\bibitem[Bertinetto et~al.(2019)Bertinetto, Henriques, Torr, and
  Vedaldi]{bertinetto2018meta}
Luca Bertinetto, Joao~F Henriques, Philip~HS Torr, and Andrea Vedaldi.
\newblock Meta-learning with differentiable closed-form solvers.
\newblock \emph{In Proceedings of the International Conference on Learning
  Representations, {ICLR}}, 2019.

\bibitem[Chang et~al.(2020)Chang, Zhang, Yu, and Jaakkola]{chang2020invariant}
Shiyu Chang, Yang Zhang, Mo~Yu, and Tommi Jaakkola.
\newblock Invariant rationalization.
\newblock In \emph{International Conference on Machine Learning, {ICML}}, pp.\
  1448--1458. PMLR, 2020.

\bibitem[Chen et~al.(2020)Chen, Kornblith, Norouzi, and Hinton]{chen2020simple}
Ting Chen, Simon Kornblith, Mohammad Norouzi, and Geoffrey Hinton.
\newblock A simple framework for contrastive learning of visual
  representations.
\newblock In \emph{International Conference on Machine Learning, {ICML}}, pp.\
  1597--1607. PMLR, 2020.

\bibitem[Chen et~al.(2019)Chen, Liu, Kira, Wang, and Huang]{ChenLKWH19}
Wei{-}Yu Chen, Yen{-}Cheng Liu, Zsolt Kira, Yu{-}Chiang~Frank Wang, and
  Jia{-}Bin Huang.
\newblock A closer look at few-shot classification.
\newblock In \emph{International Conference on Learning Representations,
  {ICLR}}, 2019.

\bibitem[Chen et~al.(2021)Chen, Liu, Xu, Darrell, and Wang]{chen2021meta}
Yinbo Chen, Zhuang Liu, Huijuan Xu, Trevor Darrell, and Xiaolong Wang.
\newblock Meta-baseline: exploring simple meta-learning for few-shot learning.
\newblock In \emph{Proceedings of the IEEE/CVF International Conference on
  Computer Vision, {ICCV}}, 2021.

\bibitem[Cheng et~al.(2023)Cheng, Yang, Zhou, Guo, and Wen]{cheng2023frequency}
Hao Cheng, Siyuan Yang, Joey~Tianyi Zhou, Lanqing Guo, and Bihan Wen.
\newblock Frequency guidance matters in few-shot learning.
\newblock In \emph{Proceedings of the IEEE/CVF International Conference on
  Computer Vision, {ICCV}}, pp.\  11814--11824, 2023.

\bibitem[Cimpoi et~al.(2014)Cimpoi, Maji, Kokkinos, Mohamed, and
  Vedaldi]{cimpoi2014describing}
Mircea Cimpoi, Subhransu Maji, Iasonas Kokkinos, Sammy Mohamed, and Andrea
  Vedaldi.
\newblock Describing textures in the wild.
\newblock In \emph{Proceedings of the IEEE Conference on Computer Vision and
  Pattern Recognition, {CVPR}}, pp.\  3606--3613, 2014.

\bibitem[Codella et~al.(2019)Codella, Rotemberg, Tschandl, Celebi, Dusza,
  Gutman, Helba, Kalloo, Liopyris, Marchetti, et~al.]{codella2019skin}
Noel Codella, Veronica Rotemberg, Philipp Tschandl, M~Emre Celebi, Stephen
  Dusza, David Gutman, Brian Helba, Aadi Kalloo, Konstantinos Liopyris, Michael
  Marchetti, et~al.
\newblock Skin lesion analysis toward melanoma detection 2018: A challenge
  hosted by the international skin imaging collaboration (isic).
\newblock \emph{arXiv preprint arXiv:1902.03368}, 2019.

\bibitem[Deng et~al.(2009)Deng, Dong, Socher, Li, Li, and
  Fei-Fei]{deng2009imagenet}
Jia Deng, Wei Dong, Richard Socher, Li-Jia Li, Kai Li, and Li~Fei-Fei.
\newblock Imagenet: A large-scale hierarchical image database.
\newblock In \emph{2009 IEEE Conference on Computer Vision and Pattern
  Recognition, {CVPR}}, pp.\  248--255. Ieee, 2009.

\bibitem[Doersch et~al.(2020)Doersch, Gupta, and
  Zisserman]{doersch2020crosstransformers}
Carl Doersch, Ankush Gupta, and Andrew Zisserman.
\newblock Crosstransformers: spatially-aware few-shot transfer.
\newblock In \emph{Advances in neural information processing systems,
  {NeurIPS}}, volume~33, pp.\  21981--21993, 2020.

\bibitem[Finn et~al.(2017)Finn, Abbeel, and Levine]{FinnAL17}
Chelsea Finn, Pieter Abbeel, and Sergey Levine.
\newblock Model-agnostic meta-learning for fast adaptation of deep networks.
\newblock In \emph{Proceedings of the 34th International Conference on Machine
  Learning, {ICML}}, 2017.

\bibitem[Gidaris et~al.(2019)Gidaris, Bursuc, Komodakis, P{\'e}rez, and
  Cord]{gidaris2019boosting}
Spyros Gidaris, Andrei Bursuc, Nikos Komodakis, Patrick P{\'e}rez, and Matthieu
  Cord.
\newblock Boosting few-shot visual learning with self-supervision.
\newblock In \emph{Proceedings of the IEEE/CVF international conference on
  computer vision, {ICCV}}, pp.\  8059--8068, 2019.

\bibitem[Gordon et~al.(2018)Gordon, Bronskill, Bauer, Nowozin, and
  Turner]{gordon2018versa}
Jonathan Gordon, John Bronskill, Matthias Bauer, Sebastian Nowozin, and
  Richard~E Turner.
\newblock Versa: Versatile and efficient few-shot learning.
\newblock In \emph{Third workshop on Bayesian Deep Learning}, 2018.

\bibitem[Guo et~al.(2020)Guo, Codella, Karlinsky, Codella, Smith, Saenko,
  Rosing, and Feris]{GuoCKCSSRF20}
Yunhui Guo, Noel Codella, Leonid Karlinsky, James~V. Codella, John~R. Smith,
  Kate Saenko, Tajana Rosing, and Rog{\'{e}}rio Feris.
\newblock A broader study of cross-domain few-shot learning.
\newblock In \emph{European Conference on Computer Vision, {ECCV}}, 2020.

\bibitem[He et~al.(2020)He, Fan, Wu, Xie, and Girshick]{he2020momentum}
Kaiming He, Haoqi Fan, Yuxin Wu, Saining Xie, and Ross Girshick.
\newblock Momentum contrast for unsupervised visual representation learning.
\newblock In \emph{Proceedings of the IEEE/CVF Conference on Computer Vision
  and Pattern Recognition, {CVPR}}, pp.\  9729--9738, 2020.

\bibitem[He et~al.(2021)He, Shen, and Cui]{he2021towards}
Yue He, Zheyan Shen, and Peng Cui.
\newblock Towards non-iid image classification: A dataset and baselines.
\newblock \emph{Pattern Recognition}, 110:\penalty0 107383, 2021.

\bibitem[Helber et~al.(2019)Helber, Bischke, Dengel, and
  Borth]{helber2019eurosat}
Patrick Helber, Benjamin Bischke, Andreas Dengel, and Damian Borth.
\newblock Eurosat: A novel dataset and deep learning benchmark for land use and
  land cover classification.
\newblock \emph{IEEE Journal of Selected Topics in Applied Earth Observations
  and Remote Sensing}, 12\penalty0 (7):\penalty0 2217--2226, 2019.

\bibitem[Hou et~al.(2019)Hou, Chang, Ma, Shan, and Chen]{hou2019cross}
Ruibing Hou, Hong Chang, Bingpeng Ma, Shiguang Shan, and Xilin Chen.
\newblock Cross attention network for few-shot classification.
\newblock In \emph{Advances in neural information processing systems,
  {NeurIPS}}, volume~32, 2019.

\bibitem[Houben et~al.(2013)Houben, Stallkamp, Salmen, Schlipsing, and
  Igel]{houben2013detection}
Sebastian Houben, Johannes Stallkamp, Jan Salmen, Marc Schlipsing, and
  Christian Igel.
\newblock Detection of traffic signs in real-world images: The german traffic
  sign detection benchmark.
\newblock In \emph{The 2013 international joint conference on neural networks
  (IJCNN)}, pp.\  1--8. Ieee, 2013.

\bibitem[Hu et~al.(2020)Hu, Moreno, Xiao, Shen, Obozinski, Lawrence, and
  Damianou]{hu2020empirical}
Shell~Xu Hu, Pablo~G Moreno, Yang Xiao, Xi~Shen, Guillaume Obozinski, Neil~D
  Lawrence, and Andreas Damianou.
\newblock Empirical bayes transductive meta-learning with synthetic gradients.
\newblock \emph{International Conference on Learning Representations, {ICLR}},
  2020.

\bibitem[Huang et~al.(2023)Huang, Li, Li, Zheng, and Liu]{huang2023pareto}
Shanshan Huang, Haoxuan Li, Qingsong Li, Chunyuan Zheng, and Li~Liu.
\newblock Pareto invariant representation learning for multimedia
  recommendation.
\newblock In \emph{ACM International Conference on Multimedia}, 2023.

\bibitem[Jongejan et~al.(2016)Jongejan, Rowley, Kawashima, Kim, and
  Fox-Gieg]{jongejan2016quick}
Jonas Jongejan, Henry Rowley, Takashi Kawashima, Jongmin Kim, and Nick
  Fox-Gieg.
\newblock The quick, draw!-ai experiment.
\newblock \emph{Mount View, CA, accessed Feb}, 17\penalty0 (2018):\penalty0 4,
  2016.

\bibitem[Kamath et~al.(2021)Kamath, Tangella, Sutherland, and
  Srebro]{kamath2021does}
Pritish Kamath, Akilesh Tangella, Danica Sutherland, and Nathan Srebro.
\newblock Does invariant risk minimization capture invariance?
\newblock In \emph{International Conference on Artificial Intelligence and
  Statistics, {AISTATS}}, pp.\  4069--4077. PMLR, 2021.

\bibitem[Kang et~al.(2021)Kang, Kwon, Min, and Cho]{kang2021relational}
Dahyun Kang, Heeseung Kwon, Juhong Min, and Minsu Cho.
\newblock Relational embedding for few-shot classification.
\newblock In \emph{Proceedings of the IEEE/CVF International Conference on
  Computer Vision, {ICCV}}, pp.\  8822--8833, 2021.

\bibitem[Ke et~al.(2023)Ke, Cao, Ling, and Zhou]{ke2023revisiting}
Tianjun Ke, Haoqun Cao, Zenan Ling, and Feng Zhou.
\newblock Revisiting logistic-softmax likelihood in bayesian meta-learning for
  few-shot classification.
\newblock In \emph{Advances in neural information processing systems,
  {NeurIPS}}, 2023.

\bibitem[Khosla et~al.(2011)Khosla, Jayadevaprakash, Yao, and
  Li]{khosla2011novel}
Aditya Khosla, Nityananda Jayadevaprakash, Bangpeng Yao, and Fei-Fei Li.
\newblock Novel dataset for fine-grained image categorization: Stanford dogs.
\newblock In \emph{Proc. CVPR Workshop on Fine-Grained Visual Categorization,
  FGVC}, 2011.

\bibitem[Krause et~al.(2013)Krause, Stark, Deng, and Fei-Fei]{krause20133d}
Jonathan Krause, Michael Stark, Jia Deng, and Li~Fei-Fei.
\newblock 3d object representations for fine-grained categorization.
\newblock In \emph{Proceedings of the IEEE international conference on computer
  vision workshops}, pp.\  554--561, 2013.

\bibitem[Krizhevsky et~al.(2009)Krizhevsky, Hinton,
  et~al.]{krizhevsky2009learning}
Alex Krizhevsky, Geoffrey Hinton, et~al.
\newblock Learning multiple layers of features from tiny images.
\newblock 2009.

\bibitem[Krueger et~al.(2021)Krueger, Caballero, Jacobsen, Zhang, Binas, Zhang,
  Le~Priol, and Courville]{krueger2021out}
David Krueger, Ethan Caballero, Joern-Henrik Jacobsen, Amy Zhang, Jonathan
  Binas, Dinghuai Zhang, Remi Le~Priol, and Aaron Courville.
\newblock Out-of-distribution generalization via risk extrapolation (rex).
\newblock In \emph{International Conference on Machine Learning, {ICML}}, 2021.

\bibitem[Kuang et~al.(2018)Kuang, Cui, Athey, Xiong, and Li]{kuang2018stable}
Kun Kuang, Peng Cui, Susan Athey, Ruoxuan Xiong, and Bo~Li.
\newblock Stable prediction across unknown environments.
\newblock In \emph{proceedings of the 24th ACM SIGKDD international conference
  on knowledge discovery \& data mining, {KDD}}, pp.\  1617--1626, 2018.

\bibitem[Lake et~al.(2015)Lake, Salakhutdinov, and Tenenbaum]{lake2015human}
Brenden~M Lake, Ruslan Salakhutdinov, and Joshua~B Tenenbaum.
\newblock Human-level concept learning through probabilistic program induction.
\newblock \emph{Science}, 350\penalty0 (6266):\penalty0 1332--1338, 2015.

\bibitem[Lee et~al.(2020)Lee, Hwang, and Shin]{lee2020self}
Hankook Lee, Sung~Ju Hwang, and Jinwoo Shin.
\newblock Self-supervised label augmentation via input transformations.
\newblock In \emph{International Conference on Machine Learning, {ICML}}, pp.\
  5714--5724. PMLR, 2020.

\bibitem[Li et~al.(2023{\natexlab{a}})Li, Lyu, Zheng, and Wu]{li2022tdr}
Haoxuan Li, Yan Lyu, Chunyuan Zheng, and Peng Wu.
\newblock {TDR}-{CL}: Targeted doubly robust collaborative learning for
  debiased recommendations.
\newblock In \emph{International Conference on Learning Representations,
  {ICLR}}, 2023{\natexlab{a}}.

\bibitem[Li et~al.(2023{\natexlab{b}})Li, Xiao, Zheng, and Wu]{li2023balancing}
Haoxuan Li, Yanghao Xiao, Chunyuan Zheng, and Peng Wu.
\newblock Balancing unobserved confounding with a few unbiased ratings in
  debiased recommendations.
\newblock In \emph{Proceedings of the ACM Web Conference, {WWW}}, pp.\
  1305--1313, 2023{\natexlab{b}}.

\bibitem[Li et~al.(2023{\natexlab{c}})Li, Zheng, and Wu]{li2022stabledr}
Haoxuan Li, Chunyuan Zheng, and Peng Wu.
\newblock Stable{DR}: Stabilized doubly robust learning for recommendation on
  data missing not at random.
\newblock In \emph{International Conference on Learning Representations,
  {ICLR}}, 2023{\natexlab{c}}.

\bibitem[Li et~al.(2024)Li, Zheng, Xiao, Wu, Geng, Chen, and Cui]{li2024kernel}
Haoxuan Li, Chunyuan Zheng, Yanghao Xiao, Peng Wu, Zhi Geng, Xu~Chen, and Peng
  Cui.
\newblock Debiased collaborative filtering with kernel-based causal balancing.
\newblock In \emph{International Conference on Learning Representations,
  {ICLR}}, 2024.

\bibitem[Li et~al.(2022)Li, Liu, and Bilen]{li2022cross}
Wei-Hong Li, Xialei Liu, and Hakan Bilen.
\newblock Cross-domain few-shot learning with task-specific adapters.
\newblock In \emph{Proceedings of the IEEE/CVF Conference on Computer Vision
  and Pattern Recognition, {CVPR}}, pp.\  7161--7170, 2022.

\bibitem[Li et~al.(2019{\natexlab{a}})Li, Wang, Xu, Huo, Gao, and
  Luo]{li2019revisiting}
Wenbin Li, Lei Wang, Jinglin Xu, Jing Huo, Yang Gao, and Jiebo Luo.
\newblock Revisiting local descriptor based image-to-class measure for few-shot
  learning.
\newblock In \emph{Proceedings of the IEEE/CVF Conference on Computer Vision
  and Pattern Recognition, {CVPR}}, 2019{\natexlab{a}}.

\bibitem[Li et~al.(2019{\natexlab{b}})Li, Xu, Huo, Wang, Gao, and
  Luo]{li2019distribution}
Wenbin Li, Jinglin Xu, Jing Huo, Lei Wang, Yang Gao, and Jiebo Luo.
\newblock Distribution consistency based covariance metric networks for
  few-shot learning.
\newblock In \emph{Proceedings of the AAAI conference on artificial
  intelligence, {AAAI}}, pp.\  8642--8649, 2019{\natexlab{b}}.

\bibitem[Li et~al.(2023{\natexlab{d}})Li, Dong, Tian, Qin, Yang, Wang, Huo,
  Shi, Wang, Gao, et~al.]{li2021libfewshot}
Wenbin Li, Chuanqi Dong, Pinzhuo Tian, Tiexin Qin, Xuesong Yang, Ziyi Wang,
  Jing Huo, Yinghuan Shi, Lei Wang, Yang Gao, et~al.
\newblock Libfewshot: A comprehensive library for few-shot learning.
\newblock \emph{IEEE Transactions on Pattern Analysis and Machine Intelligence,
  {TPAMI}}, 2023{\natexlab{d}}.

\bibitem[Liang et~al.(2021)Liang, Zhang, Dai, and Lu]{liang2021boosting}
Hanwen Liang, Qiong Zhang, Peng Dai, and Juwei Lu.
\newblock Boosting the generalization capability in cross-domain few-shot
  learning via noise-enhanced supervised autoencoder.
\newblock In \emph{Proceedings of the IEEE/CVF International Conference on
  Computer Vision, {ICCV}}, pp.\  9424--9434, 2021.

\bibitem[Liang \& Zou(2022)Liang and Zou]{liang2022metashift}
Weixin Liang and James Zou.
\newblock Metashift: A dataset of datasets for evaluating contextual
  distribution shifts and training conflicts.
\newblock In \emph{International Conference on Learning Representations,
  {ICLR}}, 2022.

\bibitem[Lin et~al.(2014)Lin, Maire, Belongie, Hays, Perona, Ramanan,
  Doll{\'a}r, and Zitnick]{lin2014microsoft}
Tsung-Yi Lin, Michael Maire, Serge Belongie, James Hays, Pietro Perona, Deva
  Ramanan, Piotr Doll{\'a}r, and C~Lawrence Zitnick.
\newblock Microsoft coco: Common objects in context.
\newblock In \emph{13th European Conference, Zurich, Switzerland, September
  6-12, 2014, Proceedings, Part V 13, {ECCV}}, pp.\  740--755. Springer, 2014.

\bibitem[Liu et~al.(2020)Liu, Cao, Lin, Li, Zhang, Long, and
  Hu]{liu2020negative}
Bin Liu, Yue Cao, Yutong Lin, Qi~Li, Zheng Zhang, Mingsheng Long, and Han Hu.
\newblock Negative margin matters: Understanding margin in few-shot
  classification.
\newblock In \emph{Computer Vision--ECCV 2020: 16th European Conference,
  Glasgow, UK, August 23--28, 2020, Proceedings, Part IV 16}, pp.\  438--455.
  Springer, 2020.

\bibitem[Liu et~al.(2018)Liu, Lee, Park, Kim, Yang, Hwang, and
  Yang]{liu2018learning}
Yanbin Liu, Juho Lee, Minseop Park, Saehoon Kim, Eunho Yang, Sung~Ju Hwang, and
  Yi~Yang.
\newblock Learning to propagate labels: Transductive propagation network for
  few-shot learning.
\newblock \emph{arXiv preprint arXiv:1805.10002}, 2018.

\bibitem[Liu et~al.(2021)Liu, Lee, Zhu, Chen, Shi, and Yang]{liu2021multi}
Yanbin Liu, Juho Lee, Linchao Zhu, Ling Chen, Humphrey Shi, and Yi~Yang.
\newblock A multi-mode modulator for multi-domain few-shot classification.
\newblock In \emph{Proceedings of the IEEE/CVF International Conference on
  Computer Vision, {ICCV}}, pp.\  8453--8462, 2021.

\bibitem[Luo et~al.(2021)Luo, Wei, Wen, Yang, Xie, Xu, and
  Tian]{luo2021rectifying}
Xu~Luo, Longhui Wei, Liangjian Wen, Jinrong Yang, Lingxi Xie, Zenglin Xu, and
  Qi~Tian.
\newblock Rectifying the shortcut learning of background for few-shot learning.
\newblock In \emph{Advances in neural information processing systems,
  {NeurIPS}}, 2021.

\bibitem[Miller(1995)]{miller1995wordnet}
George~A Miller.
\newblock Wordnet: a lexical database for english.
\newblock \emph{Communications of the ACM}, 38\penalty0 (11):\penalty0 39--41,
  1995.

\bibitem[Mohanty et~al.(2016)Mohanty, Hughes, and
  Salath{\'e}]{mohanty2016using}
Sharada~P Mohanty, David~P Hughes, and Marcel Salath{\'e}.
\newblock Using deep learning for image-based plant disease detection.
\newblock \emph{Frontiers in plant science}, 7:\penalty0 1419, 2016.

\bibitem[Motiian et~al.(2017)Motiian, Jones, Iranmanesh, and
  Doretto]{motiian2017few}
Saeid Motiian, Quinn Jones, Seyed Iranmanesh, and Gianfranco Doretto.
\newblock Few-shot adversarial domain adaptation.
\newblock In \emph{Advances in neural information processing systems,
  {NeurIPS}}, volume~30, 2017.

\bibitem[Nilsback \& Zisserman(2008)Nilsback and
  Zisserman]{nilsback2008automated}
Maria-Elena Nilsback and Andrew Zisserman.
\newblock Automated flower classification over a large number of classes.
\newblock In \emph{2008 Sixth Indian Conference on Computer Vision, Graphics \&
  Image Processing}, pp.\  722--729. IEEE, 2008.

\bibitem[Oh et~al.(2020)Oh, Yoo, Kim, and Yun]{oh2020boil}
Jaehoon Oh, Hyungjun Yoo, ChangHwan Kim, and Se-Young Yun.
\newblock Boil: Towards representation change for few-shot learning.
\newblock In \emph{International Conference on Learning Representations,
  {ICLR}}, 2020.

\bibitem[Oh et~al.(2022)Oh, Kim, Ho, Kim, Song, and Yun]{oh2022understanding}
Jaehoon Oh, Sungnyun Kim, Namgyu Ho, Jin-Hwa Kim, Hwanjun Song, and Se-Young
  Yun.
\newblock Understanding cross-domain few-shot learning based on domain
  similarity and few-shot difficulty.
\newblock In \emph{Advances in Neural Information Processing Systems,
  {NeurIPS}}, 2022.

\bibitem[Peng et~al.(2019)Peng, Bai, Xia, Huang, Saenko, and
  Wang]{peng2019moment}
Xingchao Peng, Qinxun Bai, Xide Xia, Zijun Huang, Kate Saenko, and Bo~Wang.
\newblock Moment matching for multi-source domain adaptation.
\newblock In \emph{Proceedings of the IEEE/CVF international conference on
  computer vision}, pp.\  1406--1415, 2019.

\bibitem[Peters et~al.(2015)Peters, Buhlmann, and
  Meinshausen]{peters2015causal}
J~Peters, Peter Buhlmann, and N~Meinshausen.
\newblock Causal inference using invariant prediction: identification and
  confidence intervals. arxiv.
\newblock \emph{Methodology}, 2015.

\bibitem[Radford et~al.(2021)Radford, Kim, Hallacy, Ramesh, Goh, Agarwal,
  Sastry, Askell, Mishkin, Clark, et~al.]{radford:clip}
Alec Radford, Jong~Wook Kim, Chris Hallacy, Aditya Ramesh, Gabriel Goh,
  Sandhini Agarwal, Girish Sastry, Amanda Askell, Pamela Mishkin, Jack Clark,
  et~al.
\newblock Learning transferable visual models from natural language
  supervision.
\newblock In \emph{International Conference on Machine Learning, {ICML}}, pp.\
  8748--8763. PMLR, 2021.

\bibitem[Raghu et~al.(2020)Raghu, Raghu, Bengio, and Vinyals]{RaghuRBV20}
Aniruddh Raghu, Maithra Raghu, Samy Bengio, and Oriol Vinyals.
\newblock Rapid learning or feature reuse? towards understanding the
  effectiveness of {MAML}.
\newblock In \emph{International Conference on Learning Representations,
  {ICLR}}, 2020.

\bibitem[Rajasegaran et~al.(2020)Rajasegaran, Khan, Hayat, Khan, and
  Shah]{rajasegaran2020self}
Jathushan Rajasegaran, Salman Khan, Munawar Hayat, Fahad~Shahbaz Khan, and
  Mubarak Shah.
\newblock Self-supervised knowledge distillation for few-shot learning.
\newblock \emph{arXiv preprint arXiv:2006.09785}, 2020.

\bibitem[Rosenfeld et~al.(2020)Rosenfeld, Ravikumar, and
  Risteski]{rosenfeld2020risks}
Elan Rosenfeld, Pradeep Ravikumar, and Andrej Risteski.
\newblock The risks of invariant risk minimization.
\newblock \emph{arXiv preprint arXiv:2010.05761}, 2020.

\bibitem[Russakovsky et~al.(2015)Russakovsky, Deng, Su, Krause, Satheesh, Ma,
  Huang, Karpathy, Khosla, Bernstein, et~al.]{russakovsky2015imagenet}
Olga Russakovsky, Jia Deng, Hao Su, Jonathan Krause, Sanjeev Satheesh, Sean Ma,
  Zhiheng Huang, Andrej Karpathy, Aditya Khosla, Michael Bernstein, et~al.
\newblock Imagenet large scale visual recognition challenge.
\newblock \emph{International journal of computer vision, {IJCV}}, 2015.

\bibitem[Rusu et~al.(2019)Rusu, Rao, Sygnowski, Vinyals, Pascanu, Osindero, and
  Hadsell]{RusuRSVPOH19}
Andrei~A. Rusu, Dushyant Rao, Jakub Sygnowski, Oriol Vinyals, Razvan Pascanu,
  Simon Osindero, and Raia Hadsell.
\newblock Meta-learning with latent embedding optimization.
\newblock In \emph{International Conference on Learning Representations,
  {ICLR}}, 2019.

\bibitem[Sagawa et~al.(2019)Sagawa, Koh, Hashimoto, and Liang]{sagawa2019}
Shiori Sagawa, Pang~Wei Koh, Tatsunori~B Hashimoto, and Percy Liang.
\newblock Distributionally robust neural networks for group shifts: On the
  importance of regularization for worst-case generalization.
\newblock \emph{International Conference on Learning Representations, {ICLR}},
  2019.

\bibitem[Satorras \& Estrach(2018)Satorras and Estrach]{satorras2018few}
Victor~Garcia Satorras and Joan~Bruna Estrach.
\newblock Few-shot learning with graph neural networks.
\newblock In \emph{International Conference on Learning Representations,
  {ICLR}}, 2018.

\bibitem[Schroeder \& Cui(2018)Schroeder and Cui]{schroeder2018fgvcx}
Brigit Schroeder and Yin Cui.
\newblock Fgvcx fungi classification challenge 2018.
\newblock \emph{Available online: github. com/visipedia/fgvcx\_fungi\_comp
  (accessed on 14 July 2021)}, 2018.

\bibitem[Shen et~al.(2021)Shen, Liu, He, Zhang, Xu, Yu, and
  Cui]{shen2021towards}
Zheyan Shen, Jiashuo Liu, Yue He, Xingxuan Zhang, Renzhe Xu, Han Yu, and Peng
  Cui.
\newblock Towards out-of-distribution generalization: A survey.
\newblock \emph{arXiv preprint arXiv:2108.13624}, 2021.

\bibitem[Snell et~al.(2017)Snell, Swersky, and Zemel]{snell2017prototypical}
Jake Snell, Kevin Swersky, and Richard Zemel.
\newblock Prototypical networks for few-shot learning.
\newblock In \emph{Advances in neural information processing systems,
  {NeurIPS}}, 2017.

\bibitem[Song et~al.(2022)Song, Wang, Cai, Mondal, and
  Sahoo]{song2022comprehensive}
Yisheng Song, Ting Wang, Puyu Cai, Subrota~K Mondal, and Jyoti~Prakash Sahoo.
\newblock A comprehensive survey of few-shot learning: Evolution, applications,
  challenges, and opportunities.
\newblock \emph{ACM Computing Surveys}, 2022.

\bibitem[Su et~al.(2020)Su, Maji, and Hariharan]{su2020does}
Jong-Chyi Su, Subhransu Maji, and Bharath Hariharan.
\newblock When does self-supervision improve few-shot learning?
\newblock In \emph{European Conference on Computer Vision, {ECCV}}, pp.\
  645--666. Springer, 2020.

\bibitem[Sun et~al.(2021)Sun, Lapuschkin, Samek, Zhao, Cheung, and
  Binder]{sun2021explanation}
Jiamei Sun, Sebastian Lapuschkin, Wojciech Samek, Yunqing Zhao, Ngai-Man
  Cheung, and Alexander Binder.
\newblock Explanation-guided training for cross-domain few-shot classification.
\newblock In \emph{25th International Conference on Pattern Recognition,
  {ICPR}}, 2021.

\bibitem[Sun et~al.(2019)Sun, Liu, Chua, and Schiele]{sun2019meta}
Qianru Sun, Yaoyao Liu, Tat-Seng Chua, and Bernt Schiele.
\newblock Meta-transfer learning for few-shot learning.
\newblock In \emph{Proceedings of the IEEE/CVF Conference on Computer Vision
  and Pattern Recognition}, pp.\  403--412, 2019.

\bibitem[Sung et~al.(2018)Sung, Yang, Zhang, Xiang, Torr, and
  Hospedales]{sung2017learning}
Flood Sung, Yongxin Yang, Li~Zhang, Tao Xiang, Philip~HS Torr, and Timothy~M
  Hospedales.
\newblock Learning to compare: relation network for few-shot learning.
\newblock \emph{{IEEE} Conference on Computer Vision and Pattern Recognition,
  {CVPR}}, 2018.

\bibitem[Tang et~al.(2024)Tang, Lv, Zhang, Wu, and Kuang]{tang2024modelgpt}
Zihao Tang, Zheqi Lv, Shengyu Zhang, Fei Wu, and Kun Kuang.
\newblock Modelgpt: Unleashing llm's capabilities for tailored model
  generation, 2024.

\bibitem[Tian et~al.(2023)Tian, Feng, Chai, Chen, Wang, Liu, and
  Chen]{tian2023prototypes}
Long Tian, Jingyi Feng, Xiaoqiang Chai, Wenchao Chen, Liming Wang, Xiyang Liu,
  and Bo~Chen.
\newblock Prototypes-oriented transductive few-shot learning with conditional
  transport.
\newblock In \emph{Proceedings of the IEEE/CVF International Conference on
  Computer Vision, {ICCV}}, pp.\  16317--16326, 2023.

\bibitem[Tian et~al.(2020{\natexlab{a}})Tian, Wang, Krishnan, Tenenbaum, and
  Isola]{tian2020rethinking}
Yonglong Tian, Yue Wang, Dilip Krishnan, Joshua~B Tenenbaum, and Phillip Isola.
\newblock Rethinking few-shot image classification: a good embedding is all you
  need?
\newblock In \emph{European Conference on Computer Vision, {ECCV}},
  2020{\natexlab{a}}.

\bibitem[Tian et~al.(2020{\natexlab{b}})Tian, Zhao, Shu, Yang, Li, and
  Jia]{tian2020prior}
Zhuotao Tian, Hengshuang Zhao, Michelle Shu, Zhicheng Yang, Ruiyu Li, and Jiaya
  Jia.
\newblock Prior guided feature enrichment network for few-shot segmentation.
\newblock \emph{IEEE Transactions on Pattern Analysis and Machine Intelligence,
  {TPAMI}}, 44\penalty0 (2):\penalty0 1050--1065, 2020{\natexlab{b}}.

\bibitem[Triantafillou et~al.(2020)Triantafillou, Zhu, Dumoulin, Lamblin, Evci,
  Xu, Goroshin, Gelada, Swersky, Manzagol, et~al.]{triantafillou2019meta}
Eleni Triantafillou, Tyler Zhu, Vincent Dumoulin, Pascal Lamblin, Utku Evci,
  Kelvin Xu, Ross Goroshin, Carles Gelada, Kevin Swersky, Pierre-Antoine
  Manzagol, et~al.
\newblock Meta-{D}ataset: A dataset of datasets for learning to learn from few
  examples.
\newblock 2020.

\bibitem[Triantafillou et~al.(2021)Triantafillou, Larochelle, Zemel, and
  Dumoulin]{triantafillou2021learning}
Eleni Triantafillou, Hugo Larochelle, Richard Zemel, and Vincent Dumoulin.
\newblock Learning a universal template for few-shot dataset generalization.
\newblock In \emph{International Conference on Machine Learning, {ICML}}, pp.\
  10424--10433. PMLR, 2021.

\bibitem[Tschandl et~al.(2018)Tschandl, Rosendahl, and
  Kittler]{tschandl2018ham10000}
Philipp Tschandl, Cliff Rosendahl, and Harald Kittler.
\newblock The ham10000 dataset, a large collection of multi-source
  dermatoscopic images of common pigmented skin lesions.
\newblock \emph{Scientific data}, 5\penalty0 (1):\penalty0 1--9, 2018.

\bibitem[Tseng et~al.(2020)Tseng, Lee, Huang, and Yang]{TsengLH020}
Hung{-}Yu Tseng, Hsin{-}Ying Lee, Jia{-}Bin Huang, and Ming{-}Hsuan Yang.
\newblock Cross-domain few-shot classification via learned feature-wise
  transformation.
\newblock In \emph{International Conference on Learning Representations,
  {ICLR}}, 2020.

\bibitem[Van~Horn et~al.(2018)Van~Horn, Mac~Aodha, Song, Cui, Sun, Shepard,
  Adam, Perona, and Belongie]{van2018inaturalist}
Grant Van~Horn, Oisin Mac~Aodha, Yang Song, Yin Cui, Chen Sun, Alex Shepard,
  Hartwig Adam, Pietro Perona, and Serge Belongie.
\newblock The inaturalist species classification and detection dataset.
\newblock In \emph{Proceedings of the IEEE Conference on Computer Vision and
  Pattern Recognition, {CVPR}}, pp.\  8769--8778, 2018.

\bibitem[Vinyals et~al.(2016)Vinyals, Blundell, Lillicrap, Kavukcuoglu, and
  Wierstra]{VinyalsBLKW16}
Oriol Vinyals, Charles Blundell, Tim Lillicrap, Koray Kavukcuoglu, and Daan
  Wierstra.
\newblock Matching networks for one shot learning.
\newblock In \emph{Advances in neural information processing systems,
  {NeurIPS}}, 2016.

\bibitem[Wah et~al.(2011)Wah, Branson, Welinder, Perona, and
  Belongie]{wah2011caltech}
Catherine Wah, Steve Branson, Peter Welinder, Pietro Perona, and Serge
  Belongie.
\newblock The caltech-ucsd birds-200-2011 dataset.
\newblock 2011.

\bibitem[Wang et~al.(2024)Wang, Fan, Chen, Li, Liu, Liu, Dai, Wang, Dong, and
  Tang]{wang2024optimal}
Hao Wang, Jiajun Fan, Zhichao Chen, Haoxuan Li, Weiming Liu, Tianqiao Liu,
  Quanyu Dai, Yichao Wang, Zhenhua Dong, and Ruiming Tang.
\newblock Optimal transport for treatment effect estimation.
\newblock In \emph{Advances in neural information processing systems,
  {NeurIPS}}, volume~36, 2024.

\bibitem[Wang \& Deng(2021)Wang and Deng]{WangD21}
Haoqing Wang and Zhi{-}Hong Deng.
\newblock Cross-domain few-shot classification via adversarial task
  augmentation.
\newblock In \emph{Proceedings of the Thirtieth International Joint Conference
  on Artificial Intelligence, {IJCAI}}, 2021.

\bibitem[Wang et~al.(2022)Wang, Yue, Ye, He, Li, and Li]{wang2022revisit}
Heng Wang, Tan Yue, Xiang Ye, Zihang He, Bohan Li, and Yong Li.
\newblock Revisit finetuning strategy for few-shot learning to transfer the
  emdeddings.
\newblock In \emph{International Conference on Learning Representations,
  {ICLR}}, 2022.

\bibitem[Wang et~al.(2017{\natexlab{a}})Wang, Cui, Wang, Pei, Zhu, and
  Yang]{wang2017community}
Xiao Wang, Peng Cui, Jing Wang, Jian Pei, Wenwu Zhu, and Shiqiang Yang.
\newblock Community preserving network embedding.
\newblock In \emph{Proceedings of the AAAI conference on artificial
  intelligence, {AAAI}}, volume~31, 2017{\natexlab{a}}.

\bibitem[Wang et~al.(2017{\natexlab{b}})Wang, Peng, Lu, Lu, Bagheri, and
  Summers]{wang2017chestx}
Xiaosong Wang, Yifan Peng, Le~Lu, Zhiyong Lu, Mohammadhadi Bagheri, and
  Ronald~M Summers.
\newblock Chestx-ray8: Hospital-scale chest x-ray database and benchmarks on
  weakly-supervised classification and localization of common thorax diseases.
\newblock In \emph{Proceedings of the IEEE Conference on Computer Vision and
  Pattern Recognition, {CVPR}}, pp.\  2097--2106, 2017{\natexlab{b}}.

\bibitem[Wertheimer et~al.(2021)Wertheimer, Tang, and
  Hariharan]{wertheimer2021few}
Davis Wertheimer, Luming Tang, and Bharath Hariharan.
\newblock Few-shot classification with feature map reconstruction networks.
\newblock In \emph{Proceedings of the IEEE/CVF Conference on Computer Vision
  and Pattern Recognition, {CVPR}}, pp.\  8012--8021, 2021.

\bibitem[Wu et~al.(2022)Wu, Li, Deng, Hu, Dai, Dong, Sun, Zhang, and
  Zhou]{Wu-etal2022-framework}
Peng Wu, Haoxuan Li, Yuhao Deng, Wenjie Hu, Quanyu Dai, Zhenhua Dong, Jie Sun,
  Rui Zhang, and Xiao-Hua Zhou.
\newblock On the opportunity of causal learning in recommendation systems:
  Foundation, estimation, prediction and challenges.
\newblock In \emph{International Joint Conference on Artificial Intelligence,
  {IJCAI}}, 2022.

\bibitem[Wu et~al.(2020)Wu, Nethery, Sabath, Braun, and
  Dominici]{wu2020exposure}
Xiao Wu, Rachel~C Nethery, M~Benjamin Sabath, Danielle Braun, and Francesca
  Dominici.
\newblock Exposure to air pollution and covid-19 mortality in the united
  states: A nationwide cross-sectional study.
\newblock \emph{MedRxiv}, pp.\  2020--04, 2020.

\bibitem[Xie et~al.(2022)Xie, Long, Lv, Wang, and Li]{xie2022joint}
Jiangtao Xie, Fei Long, Jiaming Lv, Qilong Wang, and Peihua Li.
\newblock Joint distribution matters: Deep brownian distance covariance for
  few-shot classification.
\newblock In \emph{Proceedings of the IEEE/CVF Conference on Computer Vision
  and Pattern Recognition, {CVPR}}, pp.\  7972--7981, 2022.

\bibitem[Yang et~al.(2022)Yang, Wang, and Zhu]{yang2022few}
Zhanyuan Yang, Jinghua Wang, and Yingying Zhu.
\newblock Few-shot classification with contrastive learning.
\newblock In \emph{European Conference on Computer Vision, {ECCV}}, pp.\
  293--309. Springer, 2022.

\bibitem[Yao et~al.(2022)Yao, Wang, Li, Zhang, Liang, Zou, and
  Finn]{yao2022improving}
Huaxiu Yao, Yu~Wang, Sai Li, Linjun Zhang, Weixin Liang, James Zou, and Chelsea
  Finn.
\newblock Improving out-of-distribution robustness via selective augmentation.
\newblock In \emph{International Conference on Machine Learning, {ICML}}, pp.\
  25407--25437. PMLR, 2022.

\bibitem[Yue et~al.(2020)Yue, Zhang, Sun, and Hua]{YueZS020}
Zhongqi Yue, Hanwang Zhang, Qianru Sun, and Xian{-}Sheng Hua.
\newblock Interventional few-shot learning.
\newblock 2020.

\bibitem[Zhang et~al.(2023{\natexlab{a}})Zhang, Gao, Luo, Shen, and
  Song]{zhang2023deta}
Ji~Zhang, Lianli Gao, Xu~Luo, Hengtao Shen, and Jingkuan Song.
\newblock Deta: Denoised task adaptation for few-shot learning.
\newblock \emph{Proceedings of the IEEE/CVF International Conference on
  Computer Vision, {ICCV}}, 2023{\natexlab{a}}.

\bibitem[Zhang et~al.(2020)Zhang, Wang, and Gai]{zhang2020knowledge}
Min Zhang, Donglin Wang, and Sibo Gai.
\newblock Knowledge distillation for model-agnostic meta-learning.
\newblock In \emph{European conference on artificial intelligence, {ECAI}},
  pp.\  1355--1362. 2020.

\bibitem[Zhang et~al.(2022{\natexlab{a}})Zhang, Huang, Li, and
  Wang]{zhang2022tree}
Min Zhang, Siteng Huang, Wenbin Li, and Donglin Wang.
\newblock Tree structure-aware few-shot image classification via hierarchical
  aggregation.
\newblock In \emph{European Conference on Computer Vision, {ECCV}}, pp.\
  453--470. Springer, 2022{\natexlab{a}}.

\bibitem[Zhang et~al.(2022{\natexlab{b}})Zhang, Huang, and
  Wang]{zhang2022domain}
Min Zhang, Siteng Huang, and Donglin Wang.
\newblock Domain generalized few-shot image classification via meta
  regularization network.
\newblock In \emph{IEEE International Conference on Acoustics, Speech and
  Signal Processing, {ICASSP}}, pp.\  3748--3752. IEEE, 2022{\natexlab{b}}.

\bibitem[Zhang et~al.(2023{\natexlab{b}})Zhang, Yuan, He, Li, Chen, and
  Kuang]{zhang2023map}
Min Zhang, Junkun Yuan, Yue He, Wenbin Li, Zhengyu Chen, and Kun Kuang.
\newblock {MAP}: Towards balanced generalization of iid and ood through
  model-agnostic adapters.
\newblock In \emph{Proceedings of the IEEE/CVF International Conference on
  Computer Vision, {ICCV}}, pp.\  11921--11931, 2023{\natexlab{b}}.

\bibitem[Zhang et~al.(2023{\natexlab{c}})Zhang, Zhuang, Wang, Wang, and
  Li]{zhang2023rotogbml}
Min Zhang, Zifeng Zhuang, Zhitao Wang, Donglin Wang, and Wenbin Li.
\newblock Roto{GBML}: Towards out-of-distribution generalization for
  gradient-based meta-learning.
\newblock In \emph{IEEE International Conference on Multimedia and Expo,
  {ICME}}, 2023{\natexlab{c}}.

\bibitem[Zhao et~al.(2021)Zhao, Ding, Lu, Xiang, Niu, Guan, and
  Wen]{zhao2021domain}
An~Zhao, Mingyu Ding, Zhiwu Lu, Tao Xiang, Yulei Niu, Jiechao Guan, and Ji-Rong
  Wen.
\newblock Domain-adaptive few-shot learning.
\newblock In \emph{Proceedings of the IEEE/CVF Winter Conference on
  Applications of Computer Vision, {WACV}}, pp.\  1390--1399, 2021.

\bibitem[Zhou et~al.(2017)Zhou, Lapedriza, Khosla, Oliva, and
  Torralba]{zhou2017places}
Bolei Zhou, Agata Lapedriza, Aditya Khosla, Aude Oliva, and Antonio Torralba.
\newblock Places: A 10 million image database for scene recognition.
\newblock \emph{IEEE Transactions on Pattern Analysis and Machine Intelligence,
  {TPAMI}}, 40\penalty0 (6):\penalty0 1452--1464, 2017.

\bibitem[Zhu et~al.(2024)Zhu, Wu, Li, Xiong, Li, Yang, Qin, Zhen, Guo, Wu,
  et~al.]{zhu2024contrastive}
Minqin Zhu, Anpeng Wu, Haoxuan Li, Ruoxuan Xiong, Bo~Li, Xiaoqing Yang, Xuan
  Qin, Peng Zhen, Jiecheng Guo, Fei Wu, et~al.
\newblock Contrastive balancing representation learning for heterogeneous
  dose-response curves estimation.
\newblock In \emph{Proceedings of the AAAI conference on artificial
  intelligence, {AAAI}}, volume~38, pp.\  17175--17183, 2024.

\end{thebibliography}
\bibliographystyle{iclr2024_conference}

\clearpage
% 不同类别不同上下文信息展示
% 图片信息展示

\appendix
\section{More Discussion on the Existing Benchmarks}\label{app-a}

In Table~\ref{tab:data}, we have summarized statistics of existing benchmarks. A brief introduction of benchmarks mentioned in this paper is the following. For more details, please refer to the original paper.  

\textbf{\textit{mini}ImageNet~\citep{VinyalsBLKW16}.}
\textit{mini}ImageNet is the subsets of the ILSVRC-12 dataset~\citep{russakovsky2015imagenet}.

\textbf{Fine-grained benchmarks.}
CUB-200-2011~\citep{wah2011caltech}, Stanford Dogs~\citep{khosla2011novel} and Stanford Cars~\citep{krause20133d} are initially designed for fine-grained classification.  

\textbf{Meta-dataset~\citep{triantafillou2019meta}.} 
Meta-dataset is a cross-domain FSC benchmark and has \textit{10 existing datasets}, including ILSVRC-2012~\citep{deng2009imagenet}, Omniglot~\citep{lake2015human}, Aircraft~\citep{wah2011caltech}, CUB-200-2011~\citep{wah2011caltech}, Describable Textures~\citep{cimpoi2014describing}, Quick Draw~\citep{jongejan2016quick}, Fungi~\citep{schroeder2018fgvcx}, VGG Flower~\citep{nilsback2008automated}, Traffic Signs~\citep{houben2013detection} and MSCOCO~\citep{lin2014microsoft}. 

\textbf{BSCD-FSL~\citep{GuoCKCSSRF20}.} 
BSCD-FSL is also a cross-domain FSC benchmark and has \textit{4 existing datasets}, including CropDiseases~\citep{mohanty2016using}, EuroSAT~\citep{helber2019eurosat}, ISIC2018~\citep{codella2019skin,tschandl2018ham10000}, and ChestX~\citep{wang2017chestx}. 
 
% \begin{figure}[tbp]
%   \centering
%   \vspace{-1mm}
%   \includegraphics[height=0.23\textwidth]{figures/similar1.pdf}
%   \includegraphics[height=0.23\textwidth]{figures/similar2.pdf}
%   \includegraphics[height=0.23\textwidth]{figures/similar3.pdf}
%   \includegraphics[height=0.23\textwidth]{figures/similar4.pdf}
%   \vspace{-2mm}
%   \caption{Similarity of each concept and image between FSC benchmarks and our MetaCoCo.}
%   \label{fig:similarity-1}
%   \vspace{-1mm}
% \end{figure}

\section{Detailed Dataset Statistics}\label{app-b}
Tables \ref{dataset_0} and \ref{dataset_1} show the number of samples for concepts and detailed statistics of the MetaCoCo benchmark, respectively. In particular, our benchmark contains 100 concepts (or categories), 155 contexts and 17.6k images. These concepts are from common objects following DomainNet~\citep{peng2019moment}. The 155 contexts are collected from the adjectives or nouns appeared more frequently with these concepts from WordNet~\citep{miller1995wordnet}. In addition, we show the statistics of samples in each concept in Table~\ref{dataset_0}. 

\section{Experimental Details}\label{app-c}

In this paper, many feature backbones are used to fair evaluate the performance of few-shot classification methods. Specifically, Conv64F contains four convolutional blocks, each of which consists of a convolutional (Conv) layer, a batch-normalization (BN) layer, a ReLU/LeakyReLU layer and a max-pooling (MP) layer, where the numbers of filters of these blocks are \{64, 64, 64, 64\}. ResNet12 consists of four residual blocks, each of which further contains three convolutional blocks (each is built as Conv-BN-ReLU-MP) along with a skip connection layer, where the numbers of filters of these blocks are \{64, 160, 320, 640\}. ResNet18 is the standard architecture used in previous works. One important difference between ResNet12 and ResNet18 is that ResNet12 uses Dropblock in each residual block, while ResNet18 does not. In addition, the number of filters of these blocks in ResNet18 is \{64, 128, 256, 512\}. ResNet10 is a common backbone architecture in cross-domain few-shot classification methods. It has the same number of filters of blocks as ResNet18, but the number of layers is $1$ in each stage. WRN-28-10 is frequently used in self-supervised learning methods, where $28$ means the number of layers and $10$ is the number of the width. 

\vspace{-3mm}
\section{More Experiments}\label{app-d}

\begin{figure}[htbp]
  \centering
  \vspace{-1mm}
  \includegraphics[height=0.22\textwidth]{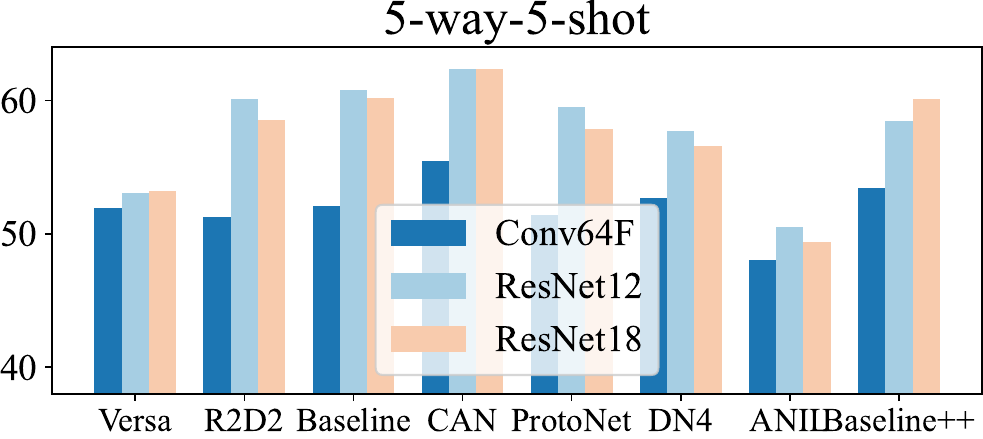}
  \includegraphics[height=0.22\textwidth]{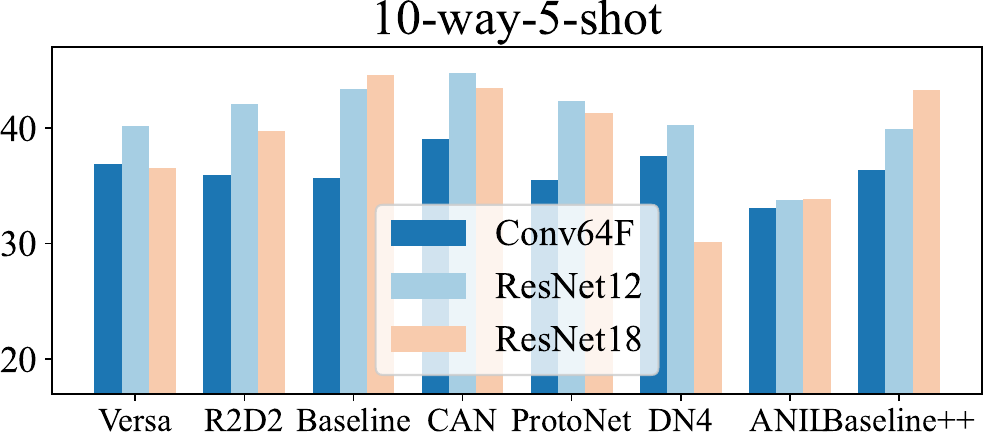}
  \vspace{-2mm}
  \caption{Experiments of different backbone architectures under 5-way and 10-way 5- shot settings.}
  \label{afig:backbone}
  \vspace{-1mm}
\end{figure}

\begin{figure}[htbp]
  \centering
  \vspace{-1mm}
  \includegraphics[height=0.19\textwidth]{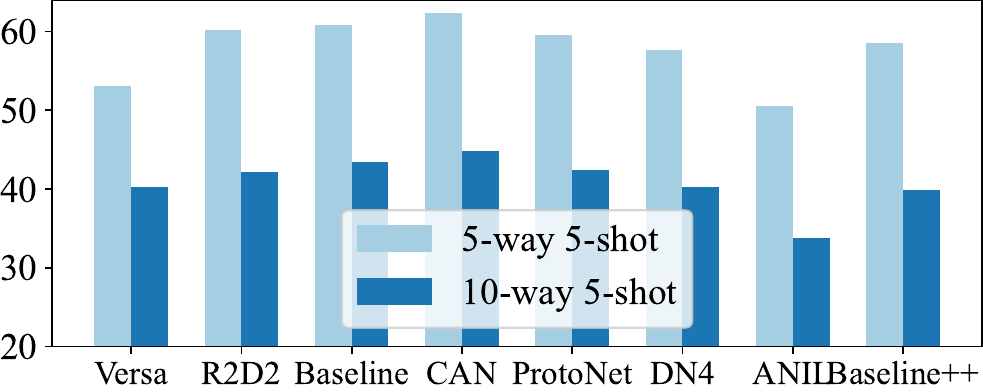}
  \includegraphics[height=0.19\textwidth]{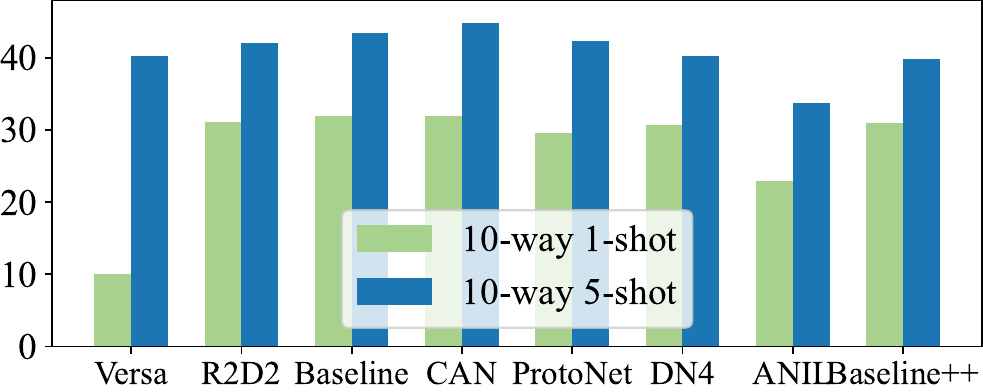}
  \vspace{-2mm}
  \caption{Experimental results of different ways (left) and shots (right) on testing performance.}
  \label{afig:way_shot}
  \vspace{-1mm}
\end{figure}

In Tables~\ref{tab:5way_acc} and~\ref{tab:10way_acc} and Figures~\ref{afig:backbone} and~\ref{afig:way_shot}, we show the additional experiments to supplement the results in our main paper. 
From these additional experiments, we find that most of the existing few-shot classification methods are not robust in the spurious-correlation problem. We hope that these studies and the proposed MetaCOCo can facilitate future research on real-world problems.

\begin{table}[htbp]
\centering
\caption{Number of samples for concepts in the MetaCoCo benchmark.}
\vspace{-3mm}
\renewcommand{\arraystretch}{1.2}
\renewcommand{\tabcolsep}{1.2mm}
\resizebox{\linewidth}{!}{
\begin{tabular}{|l c c c c c c c c c c c c c c c c c c c c c c c|}
\hline
\textcolor{blue}{Training Concepts} & dog & cat & bird & table & tree & bear & horse & fence & car & bicycle & motorcycle & train & cow & elephant & bus & chair & truck & airplane & pants & sheep & helicopter & door & monkey  \\
\textcolor{blue}{Training Samples} & 5820 & 4915 & 4397 & 4365 & 4035 & 4004 & 3980 & 3953 & 3873 & 3787 & 3755 & 3726 & 3612 & 3519 & 3491 & 3454 & 3393 & 3208 & 3080 & 3031 & 2994 & 2581 & 2578 \\
\hline
\textcolor{blue}{Training Concepts} & umbrella & lion & squirrel & boat & wolf & lizard & tiger & giraffe & tent & hot air balloon & owl & sailboat & seal & frog & jacket & rabbit & goose & kangaroo & flower & ship & cactus & hat & fox  \\
\textcolor{blue}{Training Samples} & 2499 & 2470 & 2312 & 2310 & 2293 & 2215 & 2184 & 2115 & 2038 & 2017 & 2002 & 2001 & 1976 & 1943 & 1892 & 1877 & 1862 & 1859 & 1835 & 1815 & 1801 & 1770 & 1751 \\
\hline
\textcolor{blue}{Training Concepts} & clock & spider & ostrich & tortoise & butterfly & pumpkin & sunflower & crocodile & bench & mailbox & lifeboat & dolphin & crab & window & pineapple & shorts & bag & toilet &  &  &  &  &  \\
\textcolor{blue}{Training Samples} & 1729 & 1723 & 1703 & 1672 & 1619 & 1588 & 1548 & 1499 & 1491 & 1466 & 1451 & 1182 & 1155 & 1141 & 1072 & 1026 & 993 & 864 &  &  &  &  & \\
\hline
\textcolor{blue}{Validation Concepts} & carpet & cup & refrigerator & house & zebra & tower & ocean & spoon & suit & fire hydrant & skateboard & pillow & bed & knife & backpack & bridge & & & & & & & \\
\textcolor{blue}{Validation Samples} & 437 & 434 & 431 & 430 & 420 & 381 & 379 & 362 & 357 & 347 & 340 & 329 & 317 & 314 & 312 & 249 & & & & & & &  \\
\hline
\end{tabular}
}
\label{dataset_0}
\end{table}

\begin{table}
\centering
\caption{Detailed statistics of the MetaCoCo benchmark.}
\vspace{-3mm}
\renewcommand{\arraystretch}{1.2}
\renewcommand{\tabcolsep}{1.2mm}
\resizebox{\linewidth}{!}{
\begin{tabular}{|l c c c c c c c c c c c c c c c c c c c c c c c|}
\hline
\cline{1-24}
              \multicolumn{24}{|c|}{\textcolor{blue}{Concepts}}\\
\cline{1-24}
\hline
\textcolor{blue}{Contexts} & dog & cat & bird & table & tree & bear & horse & fence & car & bicycle & motorcycle & train & cow & elephant & bus & chair & truck & airplane & pants & sheep & helicopter & door & monkey \\
\hline
grass  & 649 & 597 & 839 & 0 & 0 & 665 & 604 & 0 & 321 & 583 & 452 & 346 & 1,009 & 559 & 332 & 213 & 440 & 331 & 0 & 573 & 453 & 0 & 428  \\ 
water  & 886 & 635 & 606 & 0 & 0 & 550 & 305 & 0 & 306 & 327 & 390 & 271 & 387 & 402 & 348 & 503 & 442 & 256 & 0 & 292 & 249 & 0 & 251  \\ 
outdoor  & 456 & 249 & 203 & 0 & 0 & 131 & 375 & 0 & 402 & 415 & 347 & 202 & 263 & 286 & 367 & 276 & 246 & 345 & 0 & 395 & 436 & 0 & 231  \\ 
dim  & 322 & 241 & 414 & 0 & 0 & 164 & 318 & 0 & 390 & 521 & 489 & 333 & 268 & 338 & 202 & 145 & 486 & 299 & 0 & 303 & 477 & 0 & 184  \\ 
white  & 218 & 153 & 63 & 426 & 22 & 156 & 110 & 256 & 209 & 0 & 37 & 136 & 53 & 0 & 252 & 163 & 238 & 263 & 505 & 60 & 0 & 537 & 0  \\ 
rock  & 240 & 187 & 301 & 0 & 0 & 193 & 174 & 0 & 285 & 327 & 266 & 219 & 150 & 128 & 187 & 206 & 232 & 268 & 0 & 335 & 225 & 0 & 337  \\
\hline
\end{tabular}
}
\label{dataset_1}
\end{table}

\begin{table}
\centering
% \caption{Detailed statistics of the MetaCoCo benchmark.}
\vspace{-3mm}
\renewcommand{\arraystretch}{1.2}
\renewcommand{\tabcolsep}{1.2mm}
\resizebox{\linewidth}{!}{
\begin{tabular}{|l c c c c c c c c c c c c c c c c c c c c c c c|}
\hline
\cline{1-24}
              \multicolumn{24}{|c|}{\textcolor{blue}{Concepts}}\\
\cline{1-24}
\hline
\textcolor{blue}{Contexts} & dog & cat & bird & table & tree & bear & horse & fence & car & bicycle & motorcycle & train & cow & elephant & bus & chair & truck & airplane & pants & sheep & helicopter & door & monkey \\
\hline
open mouth  & 0 & 0 & 0 & 0 & 0 & 0 & 0 & 0 & 0 & 0 & 0 & 0 & 0 & 0 & 0 & 0 & 0 & 0 & 0 & 0 & 0 & 0 & 0  \\ 
on booth  & 0 & 0 & 0 & 0 & 0 & 0 & 0 & 0 & 112 & 0 & 0 & 0 & 0 & 0 & 0 & 0 & 0 & 0 & 0 & 0 & 0 & 0 & 0  \\ 
howling  & 0 & 0 & 0 & 0 & 0 & 0 & 0 & 0 & 0 & 0 & 0 & 0 & 0 & 0 & 0 & 0 & 0 & 0 & 0 & 0 & 0 & 0 & 0  \\ 
brick  & 0 & 0 & 0 & 0 & 0 & 0 & 0 & 0 & 0 & 0 & 0 & 0 & 0 & 0 & 0 & 0 & 0 & 0 & 0 & 0 & 0 & 0 & 0  \\ 
khaki  & 0 & 0 & 0 & 0 & 0 & 0 & 0 & 0 & 0 & 0 & 0 & 0 & 0 & 0 & 0 & 0 & 0 & 0 & 71 & 0 & 0 & 0 & 0  \\ 
at dock  & 0 & 0 & 0 & 0 & 0 & 0 & 0 & 0 & 0 & 0 & 0 & 0 & 0 & 0 & 0 & 0 & 0 & 0 & 0 & 0 & 0 & 0 & 0  \\
\hline
\end{tabular}
}
\label{dataset_2}
\end{table}

\begin{table}
\centering
% \caption{Detailed statistics of the MetaCoCo benchmark.}
\vspace{-3mm}
\renewcommand{\arraystretch}{1.2}
\renewcommand{\tabcolsep}{1.2mm}
\resizebox{\linewidth}{!}{
\begin{tabular}{|l c c c c c c c c c c c c c c c c c c c c c c c|}
\hline
\cline{1-24}
              \multicolumn{24}{|c|}{\textcolor{blue}{Concepts}}\\
\cline{1-24}
\hline
\textcolor{blue}{Contexts} & umbrella & lion & squirrel & boat & wolf & lizard & tiger & giraffe & tent & hot air balloon & owl & sailboat & seal & frog & jacket & rabbit & goose & kangaroo & flower & ship & cactus & hat & fox  \\
\hline
grass  & 460 & 437 & 0 & 238 & 371 & 293 & 444 & 389 & 269 & 376 & 332 & 290 & 351 & 0 & 457 & 391 & 336 & 419 & 202 & 318 & 285 & 398 & 283  \\ 
water  & 243 & 209 & 0 & 276 & 368 & 374 & 368 & 441 & 459 & 229 & 434 & 414 & 296 & 0 & 160 & 278 & 246 & 358 & 401 & 298 & 280 & 183 & 213  \\ 
outdoor  & 299 & 265 & 0 & 324 & 344 & 369 & 214 & 288 & 367 & 197 & 402 & 269 & 103 & 0 & 165 & 349 & 196 & 341 & 378 & 203 & 404 & 160 & 267  \\ 
dim  & 294 & 107 & 0 & 179 & 56 & 166 & 277 & 280 & 227 & 288 & 251 & 115 & 139 & 0 & 131 & 193 & 257 & 221 & 302 & 310 & 147 & 133 & 94  \\ 
white  & 0 & 116 & 138 & 76 & 0 & 65 & 33 & 74 & 0 & 107 & 0 & 95 & 0 & 146 & 131 & 94 & 37 & 20 & 0 & 0 & 82 & 91 & 147  \\ 
rock  & 253 & 241 & 0 & 265 & 344 & 201 & 149 & 250 & 254 & 123 & 236 & 272 & 258 & 0 & 122 & 86 & 110 & 322 & 203 & 205 & 73 & 151 & 84  \\
\hline
\end{tabular}
}
\label{dataset_3}
\end{table}

\begin{table}
\centering
% \caption{Detailed statistics of the MetaCoCo benchmark.}
\vspace{-3mm}
\renewcommand{\arraystretch}{1.2}
\renewcommand{\tabcolsep}{1.2mm}
\resizebox{\linewidth}{!}{
\begin{tabular}{|l c c c c c c c c c c c c c c c c c c c c c c c|}
\hline
\cline{1-24}
              \multicolumn{24}{|c|}{\textcolor{blue}{Concepts}}\\
\cline{1-24}
\hline
\textcolor{blue}{Contexts} & umbrella & lion & squirrel & boat & wolf & lizard & tiger & giraffe & tent & hot air balloon & owl & sailboat & seal & frog & jacket & rabbit & goose & kangaroo & flower & ship & cactus & hat & fox  \\
\hline
open mouth  & 0 & 0 & 0 & 0 & 0 & 0 & 0 & 0 & 0 & 0 & 0 & 0 & 0 & 0 & 0 & 0 & 0 & 0 & 0 & 0 & 0 & 0 & 0  \\ 
on booth  & 0 & 0 & 0 & 0 & 0 & 0 & 0 & 0 & 0 & 0 & 0 & 0 & 0 & 0 & 0 & 0 & 0 & 0 & 0 & 0 & 0 & 0 & 0  \\ 
howling  & 0 & 0 & 0 & 110 & 0 & 0 & 0 & 0 & 0 & 0 & 0 & 0 & 0 & 0 & 0 & 0 & 0 & 0 & 0 & 0 & 0 & 0 & 0  \\ 
brick  & 0 & 0 & 0 & 0 & 0 & 0 & 0 & 0 & 0 & 0 & 0 & 0 & 0 & 0 & 0 & 0 & 0 & 0 & 0 & 0 & 0 & 0 & 0  \\ 
khaki  & 0 & 0 & 0 & 0 & 0 & 0 & 0 & 0 & 0 & 0 & 0 & 0 & 0 & 0 & 0 & 0 & 0 & 0 & 0 & 0 & 0 & 0 & 0  \\ 
at dock  & 0 & 0 & 0 & 0 & 0 & 0 & 0 & 0 & 0 & 0 & 75 & 0 & 0 & 0 & 0 & 0 & 0 & 0 & 28 & 0 & 0 & 0 & 0  \\
\hline
\end{tabular}
}
\label{dataset_4}
\end{table}

\begin{table}
\centering
% \caption{Detailed statistics of the MetaCoCo benchmark.}
\vspace{-3mm}
\renewcommand{\arraystretch}{1.2}
\renewcommand{\tabcolsep}{1.2mm}
\resizebox{\linewidth}{!}{
\begin{tabular}{|l c c c c c c c c c c c c c c c c c c c c c c c|}
\hline
\cline{1-24}
              \multicolumn{24}{|c|}{\textcolor{blue}{Concepts}}\\
\cline{1-24}
\hline
\textcolor{blue}{Contexts} & clock & spider & ostrich & tortoise & butterfly & pumpkin & sunflower & crocodile & bench & mailbox & lifeboat & dolphin & crab & window & pineapple & shorts & bag & toilet & carpet & cup & refrigerator & house & zebra  \\
\hline
grass  & 267 & 286 & 357 & 388 & 235 & 327 & 255 & 0 & 307 & 124 & 99 & 281 & 0 & 240 & 0 & 0 & 0 & 0 & 0 & 0 & 0 & 0 & 0  \\ 
water  & 246 & 251 & 292 & 111 & 211 & 197 & 343 & 0 & 91 & 543 & 340 & 178 & 0 & 360 & 0 & 0 & 0 & 0 & 0 & 0 & 0 & 0 & 0  \\ 
outdoor  & 248 & 326 & 291 & 248 & 240 & 209 & 326 & 0 & 229 & 374 & 309 & 126 & 0 & 152 & 0 & 0 & 0 & 0 & 0 & 0 & 0 & 0 & 0  \\ 
dim  & 194 & 151 & 185 & 182 & 167 & 289 & 100 & 0 & 78 & 135 & 176 & 115 & 0 & 57 & 0 & 0 & 0 & 0 & 0 & 0 & 0 & 0 & 0  \\ 
white  & 76 & 0 & 0 & 0 & 0 & 0 & 40 & 47 & 0 & 0 & 54 & 0 & 150 & 0 & 278 & 108 & 663 & 46 & 138 & 271 & 140 & 93 & 60  \\ 
rock  & 98 & 113 & 198 & 255 & 118 & 137 & 151 & 0 & 234 & 123 & 93 & 183 & 0 & 53 & 0 & 0 & 0 & 0 & 0 & 0 & 0 & 0 & 0  \\
\hline
\end{tabular}
}
\label{dataset_5}
\end{table}

\begin{table}
\centering
% \caption{Detailed statistics of the MetaCoCo benchmark.}
\vspace{-3mm}
\renewcommand{\arraystretch}{1.2}
\renewcommand{\tabcolsep}{1.2mm}
\resizebox{\linewidth}{!}{
\begin{tabular}{|l c c c c c c c c c c c c c c c c c c c c c c c|}
\hline
\cline{1-24}
              \multicolumn{24}{|c|}{\textcolor{blue}{Concepts}}\\
\cline{1-24}
\hline
\textcolor{blue}{Contexts} & clock & spider & ostrich & tortoise & butterfly & pumpkin & sunflower & crocodile & bench & mailbox & lifeboat & dolphin & crab & window & pineapple & shorts & bag & toilet & carpet & cup & refrigerator & house & zebra  \\
\hline
open mouth  & 0 & 0 & 0 & 0 & 0 & 0 & 113 & 0 & 0 & 0 & 0 & 0 & 0 & 0 & 0 & 0 & 0 & 0 & 0 & 0 & 0 & 0 & 0  \\ 
on booth  & 0 & 0 & 0 & 0 & 0 & 0 & 0 & 0 & 0 & 0 & 0 & 0 & 0 & 0 & 0 & 0 & 0 & 0 & 0 & 0 & 0 & 0 & 0  \\ 
howling  & 0 & 0 & 0 & 0 & 0 & 0 & 0 & 0 & 0 & 0 & 0 & 0 & 0 & 0 & 0 & 0 & 0 & 0 & 0 & 0 & 0 & 0 & 0  \\ 
brick  & 0 & 0 & 0 & 0 & 0 & 0 & 0 & 0 & 0 & 0 & 0 & 0 & 0 & 0 & 0 & 0 & 0 & 0 & 0 & 0 & 51 & 0 & 57  \\ 
khaki  & 0 & 0 & 0 & 0 & 0 & 0 & 0 & 0 & 0 & 0 & 0 & 0 & 0 & 0 & 34 & 0 & 0 & 0 & 0 & 0 & 0 & 0 & 0  \\ 
at dock  & 0 & 0 & 0 & 0 & 0 & 0 & 0 & 0 & 0 & 0 & 0 & 0 & 0 & 0 & 0 & 0 & 0 & 0 & 0 & 0 & 0 & 0 & 0  \\
\hline
\end{tabular}
}
\label{dataset_6}
\end{table}

\begin{table}
\centering
% \caption{Detailed statistics of the MetaCoCo benchmark.}
\vspace{-3mm}
\renewcommand{\arraystretch}{1.2}
\renewcommand{\tabcolsep}{1.2mm}
\resizebox{\linewidth}{!}{
\begin{tabular}{|l c c c c c c c c c c c c c c c c c c c c c c c|}
\hline
\cline{1-24}
              \multicolumn{24}{|c|}{\textcolor{blue}{Concepts}}\\
\cline{1-24}
\hline
\textcolor{blue}{Contexts} & tower & ocean & spoon & suit & fire hydrant & skateboard & pillow & bed & knife & backpack & bridge & rat & laptop & sink & frame & bowl & coat & bush & cloud & cabinet & shrimp & dress & television  \\
\hline
grass  & 0 & 0 & 0 & 0 & 0 & 0 & 0 & 0 & 0 & 0 & 125 & 0 & 0 & 0 & 0 & 0 & 0 & 0 & 0 & 190 & 0 & 0 & 0  \\ 
water  & 0 & 0 & 0 & 0 & 0 & 0 & 0 & 0 & 0 & 0 & 0 & 0 & 0 & 0 & 0 & 0 & 0 & 0 & 0 & 260 & 0 & 0 & 0  \\ 
outdoor  & 0 & 0 & 0 & 0 & 0 & 0 & 0 & 0 & 0 & 0 & 0 & 0 & 0 & 0 & 0 & 0 & 0 & 0 & 0 & 44 & 0 & 0 & 0  \\ 
dim  & 0 & 0 & 0 & 0 & 0 & 0 & 0 & 0 & 0 & 0 & 0 & 0 & 0 & 0 & 0 & 0 & 0 & 0 & 0 & 7 & 0 & 0 & 0  \\ 
white  & 22 & 25 & 27 & 49 & 35 & 161 & 125 & 11 & 0 & 0 & 0 & 64 & 473 & 167 & 288 & 68 & 0 & 421 & 151 & 0 & 159 & 9 & 186  \\ 
rock  & 0 & 0 & 0 & 0 & 0 & 0 & 0 & 0 & 0 & 0 & 0 & 0 & 0 & 0 & 0 & 0 & 0 & 0 & 0 & 47 & 0 & 0 & 0  \\
\hline
\end{tabular}
}
\label{dataset_7}
\end{table}

\begin{table}
\centering
% \caption{Detailed statistics of the MetaCoCo benchmark.}
\vspace{-3mm}
\renewcommand{\arraystretch}{1.2}
\renewcommand{\tabcolsep}{1.2mm}
\resizebox{\linewidth}{!}{
\begin{tabular}{|l c c c c c c c c c c c c c c c c c c c c c c c|}
\hline
\cline{1-24}
              \multicolumn{24}{|c|}{\textcolor{blue}{Concepts}}\\
\cline{1-24}
\hline
\textcolor{blue}{Contexts} & tower & ocean & spoon & suit & fire hydrant & skateboard & pillow & bed & knife & backpack & bridge & rat & laptop & sink & frame & bowl & coat & bush & cloud & cabinet & shrimp & dress & television  \\
\hline
open mouth  & 0 & 0 & 0 & 0 & 0 & 0 & 0 & 0 & 0 & 0 & 0 & 0 & 0 & 0 & 0 & 0 & 0 & 0 & 0 & 0 & 0 & 0 & 0  \\ 
on booth  & 0 & 0 & 0 & 0 & 0 & 0 & 0 & 0 & 0 & 0 & 0 & 0 & 0 & 0 & 0 & 0 & 0 & 0 & 0 & 0 & 0 & 0 & 0  \\ 
howling  & 0 & 0 & 0 & 0 & 0 & 0 & 0 & 0 & 0 & 0 & 0 & 0 & 0 & 0 & 0 & 0 & 0 & 0 & 0 & 0 & 0 & 0 & 0  \\ 
brick  & 0 & 0 & 0 & 0 & 0 & 0 & 0 & 0 & 0 & 0 & 0 & 0 & 0 & 0 & 0 & 0 & 0 & 0 & 0 & 0 & 0 & 0 & 0  \\ 
khaki  & 0 & 0 & 0 & 0 & 0 & 0 & 0 & 0 & 0 & 0 & 0 & 0 & 0 & 0 & 0 & 0 & 0 & 0 & 0 & 0 & 0 & 0 & 0  \\ 
at dock  & 0 & 0 & 0 & 0 & 0 & 0 & 0 & 0 & 0 & 0 & 0 & 0 & 0 & 0 & 0 & 0 & 0 & 0 & 0 & 0 & 0 & 0 & 0  \\
\hline
\end{tabular}
}
\label{dataset_8}
\end{table}

\begin{table}
\centering
% \caption{Detailed statistics of the MetaCoCo benchmark.}
\vspace{-3mm}
\renewcommand{\arraystretch}{1.2}
\renewcommand{\tabcolsep}{1.2mm}
\resizebox{\linewidth}{!}{
\begin{tabular}{|l c c c c c c c c c c c c c c c c c c c c c c c|}
\hline
\cline{1-24}
              \multicolumn{24}{|c|}{\textcolor{blue}{Contexts}}\\
\cline{1-24}
\hline
\textcolor{blue}{Concepts} & grass & water & outdoor & dim & white & rock & autumn & black & brown & wood & blue & green & gray & on snow & large & eating & red & metal & lying & with people & on beach & small & in forest \\
\hline
t-shirt & 0 & 0 & 0 & 0 & 186 & 0 & 0 & 112 & 0 & 0 & 80 & 32 & 51 & 0 & 0 & 0 & 49 & 0 & 0 & 0 & 0 & 0 & 0 \\
sweater & 0 & 0 & 0 & 0 & 62 & 0 & 0 & 101 & 36 & 0 & 90 & 33 & 116 & 0 & 0 & 0 & 47 & 0 & 0 & 0 & 0 & 0 & 0  \\
surfboard  & 0 & 0 & 0 & 0 & 265 & 0 & 0 & 18 & 11 & 0 & 111 & 0 & 0 & 0 & 24 & 0 & 21 & 0 & 0 & 0 & 0 & 12 & 0 \\
tie  & 0 & 0 & 0 & 0 & 44 & 0 & 0 & 113 & 19 & 0 & 104 & 21 & 36 & 0 & 0 & 0 & 75 & 0 & 0 & 0 & 0 & 0 & 0  \\
fork  & 0 & 0 & 0 & 0 & 27 & 0 & 0 & 14 & 0 & 0 & 0 & 0 & 10 & 0 & 0 & 0 & 6 & 118 & 0 & 0 & 0 & 0 & 0  \\
couch & 0 & 0 & 0 & 0 & 81 & 0 & 0 & 41 & 160 & 0 & 46 & 28 & 43 & 0 & 0 & 0 & 38 & 0 & 0 & 0 & 0 & 0 & 0  \\
\hline
\end{tabular}
}
\label{dataset_9}
\end{table}

\begin{table}[tbp]  
    \centering
    \caption{Experiments of state-of-the-art few-shot classification methods under 5-way 1- and 5-shot setting. Three common backbones are used and the 95\% confidence intervals are displayed.}
    \vspace{-3mm}
    % \scalebox{0.95}{
    \resizebox{1.0\textwidth}{!}{
        \begin{tabular}{lcc|cc}
            \shline
            Method & Backbone & Type & $5$-way $1$-shot & $5$-way $5$-shot \\ 
            \hline
            \textbf{Versa}~\citep{gordon2018versa} & Conv64F & Meta  & 20.00 \scriptsize$\pm$ 0.00 & 51.95 \scriptsize$\pm$ 0.35 \\
            \textbf{R2D2}~\citep{bertinetto2018meta} & Conv64F & Meta  & 39.56 \scriptsize$\pm$ 0.36 & 51.29 \scriptsize$\pm$ 0.36 \\
             \textbf{ANIL}~\citep{RaghuRBV20} & Conv64F & Meta  & 37.24 \scriptsize$\pm$ 0.36 & 48.04 \scriptsize$\pm$ 0.39 \\
            \textbf{CAN}~\citep{hou2019cross} & Conv64F & Metric  & 43.02 \scriptsize$\pm$ 0.42 & 55.48 \scriptsize$\pm$ 0.40 \\
            \textbf{ProtoNet}~\citep{snell2017prototypical} & Conv64F & Metric  & 37.92 \scriptsize$\pm$ 0.35 & 51.39 \scriptsize$\pm$ 0.38 \\
            \textbf{DN4}~\citep{li2019revisiting} & Conv64F & Metric  & 38.87 \scriptsize$\pm$ 0.36 & 52.72 \scriptsize$\pm$ 0.38 \\
            \textbf{Baseline}~\citep{ChenLKWH19} & Conv64F & Fine-tuning  & 37.97 \scriptsize$\pm$ 0.32 & 52.08 \scriptsize$\pm$ 0.35 \\
\end{tabular}
    } 
\label{tab:5way_acc}
\end{table}
\begin{table}[tbp]  
    \centering
    \caption{Experiments of state-of-the-art few-shot classification methods under 10-way 1- and 5-shot setting. Three common backbones are used and the 95\% confidence intervals are displayed.}
    \vspace{-3mm}
    % \scalebox{0.95}{
    \resizebox{1.0\textwidth}{!}{
        \begin{tabular}{lrc|cc}
            \shline
            Method & Backbone & Type & $10$-way $1$-shot & $10$-way $5$-shot \\ 
            \hline
            \textbf{Versa}~\citep{gordon2018versa} & Conv64F & Meta & 10.00 \scriptsize$\pm$ 0.00 & 36.88 \scriptsize$\pm$ 0.21 \\
            \textbf{R2D2}~\citep{bertinetto2018meta} & Conv64F & Meta & 24.80 \scriptsize$\pm$ 0.20 & 35.98 \scriptsize$\pm$ 0.21 \\
            \textbf{ANIL}~\citep{RaghuRBV20} & Conv64F & Meta & 22.39 \scriptsize$\pm$ 0.19 & 33.05 \scriptsize$\pm$ 0.22 \\
            \textbf{CAN}~\citep{hou2019cross} & Conv64F & Metric & 28.19 \scriptsize$\pm$ 0.23 & 39.06 \scriptsize$\pm$ 0.23 \\
            \textbf{ProtoNet}~\citep{snell2017prototypical} & Conv64F & Metric & 23.72 \scriptsize$\pm$ 0.19 & 35.49 \scriptsize$\pm$ 0.22 \\
            \textbf{DN4}~\citep{li2019revisiting} & Conv64F & Metric & 26.26 \scriptsize$\pm$ 0.21 & 37.54 \scriptsize$\pm$ 0.22 \\
            \textbf{Baseline}~\citep{ChenLKWH19} & Conv64F & Fine-tuning & 23.71 \scriptsize$\pm$ 0.18 & 35.70 \scriptsize$\pm$ 0.21 \\
\end{tabular}
    } 
\label{tab:10way_acc}
\end{table}

\end{document}